%% file: ms.tex
\documentclass{article}
\usepackage{arxiv}

\usepackage[utf8]{inputenc}
\usepackage[pdftex]{graphicx}
\usepackage{mathtools}
\usepackage{amsmath}
\usepackage{array}
\usepackage{mathrsfs}
\usepackage{amssymb}
\usepackage{booktabs}
\usepackage{algorithm}
\usepackage{algpseudocode}
\usepackage{multirow}
\usepackage{tabularx}
\usepackage{tabulary}
\usepackage{dcolumn}
\usepackage{threeparttable}
\usepackage{bbm}
\usepackage{color, soul}
\usepackage{blindtext}
\usepackage{setspace}
\usepackage{geometry}
\usepackage{wrapfig}
\usepackage{lscape}
\usepackage{rotating}
\usepackage{epstopdf}
\usepackage{amsfonts}
\usepackage{pgf}
\usepackage[toc,page]{appendix}
\usepackage{siunitx}
\usepackage{tikz}
\usepackage{multirow}
\usepackage{standalone}
\usetikzlibrary{shapes,arrows}
\usepackage{hyperref}
\usepackage[english]{babel}
\usepackage{xcolor}
\usepackage{etex}

\hypersetup{
	colorlinks,
	linkcolor={red!50!black},
	citecolor={blue!50!black},
	urlcolor={blue!80!black}
}
\providecommand{\keywords}[1]
{
	\small	
	\textbf{\textit{Keywords:}} #1
}

\usepackage[numbers,sort&compress]{natbib}

\allowdisplaybreaks 

\title{Multivariate, Multistep Forecasting, Reconstruction and Feature Selection of Ocean Waves via Recurrent and Sequence-to-Sequence Networks}

\author{
	Mohammad Pirhooshyaran \\
	Industrial \& Systems Engineering\\
	 Lehigh University\\
	\texttt{mop216@lehigh.edu} \\
	\And
	Lawrence V.\ Snyder \\
	Industrial \& Systems Engineering\\
	 Lehigh University\\
	\texttt{lvs2@lehigh.edu} \\
}

\begin{document}
	
	\maketitle

\begin{abstract}
This article explores the concepts of ocean wave multivariate multistep forecasting, reconstruction and feature selection. We introduce recurrent neural network frameworks, integrated with Bayesian hyperparameter optimization and Elastic Net methods. We consider both short- and long-term forecasts and reconstruction, for significant wave height and output power of the ocean waves.
Sequence-to-sequence neural networks are being developed for the first time to reconstruct the missing characteristics of ocean waves based on information from nearby wave sensors.  Our results indicate that the Adam and AMSGrad optimization algorithms are the most robust ones to optimize the sequence-to-sequence network. For the case of significant wave height reconstruction, we compare the proposed methods with alternatives on a well-studied dataset. We show the superiority of the proposed methods considering several error metrics. We design a new case study based on measurement stations along the east coast of the United States and investigate the feature selection concept. Comparisons substantiate the benefit of utilizing Elastic Net. Moreover, case study results indicate that when the number of features is considerable, having deeper structures improves the performance. 
\end{abstract}
	
\section{Introduction}

The intricate and ever-changing character of irregular waves necessitates the existence of a framework to estimate wave features in advance. Bringing the energy generated by ocean waves into the consumer's portfolios requires a consistent system capable of predicting ocean wave uncertainties. Most research on ocean wave prediction has focused on model-based approaches---for example, physics-based models that attempt to reproduce the frequency spectra of ocean waves and then to predict the resulting wave energy. However, these models are difficult to generalize.

Machine learning (ML) techniques allow model-free predictions of ocean wave characteristics to become practical and computationally efficient via monitoring of the huge amount of records taken in diverse bodies of water all around the world. We introduce recurrent networks integrated with Bayesian optimization and Elastic Net techniques to forecast future wave features and reconstruct missing data simultaneously.

There is a substantial literature on SWH forecasting; see Section~\ref{sec:lit} for a review. These papers investigate short- to long-term predictions of SWH at coastal and offshore sites with different water depths. Although there are marine applications solely based on SWH estimation, in most cases SWH is used as an input to evaluate other characteristics of an ocean wave, such as energy flux. Energy flux is the expected total sum of the wave energy available at a given location, and it plays a major role in designing marine structures from at least two broad perspectives. First, by modeling the energy flux of any potential location, one may identify  remote locations where Wave Energy Converters (WEC) can be installed. Second, renewable energy of all sorts is well known to be less reliable sources with less resiliency against disruptions compared to energy produced by thermal power plants. Although wave energy is more predictable than wind or solar, it still suffers from randomness. Having a concrete framework to forecast significant wave height and ocean energy flux may directly improve the resiliency of electricity produced by this green source.

We propose a model that uses neural networks (NN) to forecast wave characteristics. Most studies that have used NNs to forecast or reconstruct wave characteristics use fully connected networks receiving inputs such as significant wave height ($H_s$) and/or average wave period ($A$) and outputs such as power ($\mathcal{P}$) and/or $H_s$. Recurrent Neural Networks (RNN), which contain natural temporal properties, have been considered in a few cases as well. The majority of these cases solely focus on predicting $H_s$. To the best of our knowledge, no studies have used sequence-to-sequence NNs to predict energy flux or reconstruct features. In this paper, we investigate the concept of multi-step-ahead univariate ($H_s$) and multivariate ($\mathcal{P}$) forecasting and feature reconstruction. Our contributions are as follows:
    \begin{itemize}
    	\item  We introduce two main networks. The first one is a simple combination of Long Short-Term Memory plus Fully Connected Layers at the end (LSTM+FCL). The LSTM part can be single- or multi-layered structures. The second network is a sequence-to-sequence (seqtoseq) \cite{sutskever2014sequence} method containing two LSTMs as its encoder and decoder. 
    	\item We expand the framework to tackle the problem of reconstructing  missing ocean wave data and feature selection based on information from their nearby buoys. We perform a detailed comparison between our models and the state-of-the-art papers on wave feature reconstruction \cite{cornejo2016significant,cornejo2018bayesian} and demonstrate that our approach outperforms those in the literature in terms of capturing missing $H_s$ information.
    	\item We modify an epoch-scheduled training scheme that is suitable for time series analysis in general as well as an elastic net concept, and we train the models based on these proposed schemes. A comparison between single- and multi-layered RNN, LSTM, and  seqtoseq methods is conducted.
    	\item We conduct Spearmint Bayesian optimization (GP-EI) to hypertune the models' parameters. Throughout the paper, we indicate the $i$th parameters that must be tuned as (*$i$*). 
    	\item We integrate elastic net concept into the model for the purpose of feature selection.             
    	\item We investigate the performance of four different optimization algorithms---SGD, RMSProp, Adam, and AMSGrad---for the seqtoseq network based on years of ocean wave data. 
    	\item We design a new dataset and a case study. The dataset is general and can be used for any feature selection and/or multivariate regression purposes. (Datasets are at the repository: \href{https://github.com/mamadpierre/NOAA-Refined-Stations}{NOAA-Refined-Stations})  
\end{itemize}  
 The remainder of this paper is structured as follows: In Section \ref{sec:lit}, we discuss the necessary background on ocean waves, and we provide a concise literature review. Then, we introduce the  LSTM+FCL and seqtoseq models and epoch-schedule training in section \ref{sec:model}. Section \ref{sec:exp} is dedicated to several comparisons, including long-term forecasts. The reconstruction framework is addressed in section \ref{sec:recon}. In section \ref{sec:feature_selection} we discuss about the feature selection with Elastic Net technique. The paper concludes in section \ref{conclu}.


\section{Background and Literature Review}
\label{sec:lit}

Statistical properties of the ocean surface suggest that for a given time and location, ocean waves may be viewed as the summation of a considerable amount of independent regular waves caused by wind sources or wave interactions \cite{hadjihosseini2014stochastic}. Therefore, the ocean surface can be modeled as a zero-mean Gaussian stochastic process, considering all these waves \cite{hadjihosseini2014stochastic,rychlik1997modelling}. Buoys that contain wave-measurement sensors take a large number of samples from this process over time, and then, one may recover the spectral wave density $\mathcal{S}(w)$, which is the fast Fourier transform of the covariance matrix of the sea surface elevation \cite{tucker2001waves,falnes2007review,steele1993ndbc}.

Once $\mathcal{S}(w)$ of a surface wave has been estimated from the measurements collected, the Spectral Density Momentum (SDM) of order $r$ is calculated as:	
\begin{equation} \label{momentum}
\text{SDM}(r)= \int_{0}^{\infty} w^r \mathcal{S}(w)dw.
\end{equation} 
By definition, we have $H_s =4\int_{0}^{\infty} \mathcal{S}(w)dw = 4 \text{SDM}(0)^\frac{1}{2}$ and Average wave period $A = \frac{\text{SDM}(-1)}{\text{SDM}(0)}$. The energy flux $\mathcal{P}$ is then calculated as follows: 
 \begin{equation}\label{power}
\mathcal{P}= \frac{\rho g^2}{64\pi} \int_{0}^{\infty} \frac{S(w)}{w} dw =  \frac{\rho g^2}{64\pi} (H_s)^2 A, 
\end{equation}
where $ \rho$ is the ocean density, and $g$ is the gravity constant.

The rest of this section contains three subsections. First, we provide a concise review of the literature about forecasting ocean wave characteristics. Second, we explore the novel work on oceanic wave feature reconstruction. Third, we discuss seqtoseq models from a machine learning point of view.
\subsection{Ocean Characteristic Forecasting}
Wave forecasting models can be categorized into model-based and model-free frameworks. Model-based approaches aim to use physical concepts such as climatic pressure, frictional dissipation and environmental interactions to find precise equations mimicking the behavior of wind waves. These methods are further classified based on their efforts to explicitly parameterize ocean wave  interactions \cite{tolman1991third} into first, second and third generations. First-generation models try to construct the spectral wave structure solely based on the linear wave interactions. These models overestimate the power generated by water because of linear simplification and instead ignore any nonlinear transfer \cite{group1988wam}.
Second-generation models such as JONSWAP \cite{hasselmann1973measurements} examine coupled discrete spectral structures in such a way that the wave nonlinearities can be parameterized \cite{perrie1989modelling}. The most mature models are third-generation wave models such as WAM and simulating waves nearshore (SWAN) \cite{booij1999third}, which consider all possible generation, dissipation and nonlinear wave-wave interactions \cite{group1988wam} along with current--wave interactions \cite{tolman1991third}.

In contrast, data-driven, model-free approaches have become more popular recently with the advent of ML techniques. Fuzzy Systems (FS) \cite{ozger2007prediction,kazeminezhad2005application,hashim2016selection}, Evolutionary Algorithms (EA) \cite{cornejo2016significant,cornejo2018bayesian}, Support Vector Machines (SVM) \cite{mahjoobi2009prediction}, and Deep Neural Networks \cite{deo2001neural,abhigna2017analysis,savitha2017regional} are so-called ``soft computing techniques,'' which focus on the data structure in order to investigate possible relations and dependencies to forecast uncertain future events. Neural Networks are among the most powerful tools to approximate almost any nonlinear and complex functional relation. Recurrent Neural Networks (RNNs), a subclass of Neural Networks, exploit their internal memories to express temporal dynamic behavior, which makes them a suitable framework for forecasting complex systems. \cite{malekmohamadi2011evaluating} explore the efficacy of several ML methods in terms of their accuracy in forecasting the significant wave height.

 The general goal of any forecasting method is to find an accurate short- or long-term forecast of the variable under study. (See \cite{azencott2019automatic,kamalzadeh2017shape} for new forecasting studies.) However, there exist ML approaches particularly designed for forecasting of wave characteristics. For more thorough discussion of the literature on ocean power and significant wave height forecasting one may refer to past reviews \cite{uihlein2016wave,zheng2016overview,zheng2017overview,jha2017renewable,jain2006neural,penalba2017mathematical,lehmann2017ocean}. \cite{hatalis2014multi} implements a Nonlinear Autoregressive (NAR) neural network to forecast exponentially smoothed ocean wave power using Irish Marine Institute data. A Minimal Resource Allocation Network (MRAN) and a Growing and Pruning Radial Basis Function (GAP-RBF) network are implemented and tested on three geographical locations by \cite{savitha2017regional}. The significance of a node in the GAP-RBF network is measured as its contribution to the network output, and the node is added or pruned accordingly.  Cascade-forward and feed-forward neural networks are implemented in \cite{mahmoodi2017data} to predict the wave power itself in the absence of spectral wave data. An ensemble of Extreme Learning Machines (ELM) is presented in \cite{kumar2017ocean} to predict the daily wave height. The authors use the previous hours' of wave heights along with features such as air to sea temperature difference, atmospheric pressure, wind speed to predict the next 6-hour wave height. Two computationally efficient supervised ML approaches are introduced by \cite{james2017machine} and compared with the SWAN model \cite{team2014swan}. An integrated numerical and ANN approach is introduced by \cite{londhe2016coupled} to predict waves 24 hours in advance at different buoys along the Indian Coastline. Nonlinear and non-stationary $H_s$s are studied in \cite{duan2016hybrid} based on integrated Empirical Model Decomposition Support Vector Regression (EMD-SVR). Forecasting of extreme events such as hurricanes is examined by Dixit et al. \cite{dixit2016prediction} via a Neuro Wavelet Technique (NWT). A recent work \cite{prahlada2015forecasting} aims at predicting $H_s$ 48 hours into the future utilizing a hybrid model combining neural network with wavelets (WLNN). As mentioned before non of these work take into consideration sequence-to-sequence networks to investigate its short to long term forecasting performance yet alone its combination with Spearmint Bayesian optimization and elastic net technique.

\subsection{Reconstruction of Ocean Characteristics}

In contrast to forecasting frameworks, in which the goal is typically to estimate wave features such as $H_s$ and $\mathcal{P}$ in the future based on historical data, reconstruction models aim to use available information about wave features to reconstruct  $H_s$, $\mathcal{P}$, or other (usually missing) features. Here, the assumption is that the model has access to up-to-date measurements of the wave features, except for the one(s) they want to reconstruct. This is commonly due to missing measurement data. Therefore, the prediction continues one step ahead (or a few steps ahead) into the future. Hence, the aim is to tackle the problem of extracting the ocean wave information of a location purely based on other available features. The framework is useful for estimating the missing data of a station using the knowledge obtained from its neighbors but any available information of the same station can be utilized as well.

The paper \cite{cornejo2016significant} and its subsequent improvement \cite{cornejo2018bayesian} are dedicated to reconstructing ocean waves based on Evolutionary Algorithms (EA). The authors address the problem primarily via a Grouping Genetic Algorithm (GGA) and Bayesian optimization Grouping Genetic Algorithm (BO-GGA). They utilize GGA and/or BO-GGA to select the wave features of nearby stations suitable to predict the desired wave feature of the location with missing data, and then obtain their predictions via simple ELM or SVM. The paper reconstructs the significant wave height of the NOAA buoy 46069 for the year 2010 solely based upon the information provided from two adjacent buoys.

\subsection{Sequence-to-Sequence Neural Networks}

Deep Neural Networks (DNN) are among the most successful tools for classification and regression. DNNs achieve state-of-the-art performance in many applications. Although a simple feed-forward DNN can be applied in many systems, they require the system to have fixed input and output size. Recurrent Neural Networks such as Long-Short-Term Memory (LSTM) networks tackle this limitation in the sense that they do not need a fixed input size \cite{hochreiter1997long}. LSTMs can observe an input sequence of arbitrary size sequentially (one time step at a time) to provide the rest of the network with a large, fixed-sized vector representing the input \cite{lipton2015critical,mobiny2018text}. Furthermore, LSTMs remember long-range feature propagation based on a sigmoid layer called ``forget gate.'' Seqtoseq models use two separate recurrent structures. They vary from basic recurrent networks in the sense that the network fully reads the input sequences before it generates any outputs. The first network is usually an LSTM which reads (encodes) the input of any size and maps them to a fixed-sized output. The second structure generally receives the fixed-sized output vector of the first LSTM and maps (decodes) them to a desirable output space \cite{sutskever2014sequence}. Having encoding and decoding as separate steps gives the model flexibility and stability when handling complex sequence structures \cite{chiu2017state}.\\
An important technique for training recurrent networks is Teacher Forcing (TF), which forces the network to observe the previous ground truth output instead of the one it already predicted. In other words, TF, keeping the network complex structure, converts any long-term prediction structures to one-step-ahead forecasts. This can greatly increase the network's ability to learn, thereby reducing its learning time \cite{williams1989learning,lamb2016professor}. On the other hand, one can argue that the new model is not solving the same problem anymore. Therefore, there often exists a large gap between the testing error and training error for the models trained by TF; that is, the model encounters a major overfitting problem. Hence, \cite{bengio2015scheduled} introduces a modification of scheduled training that captures the benefit TF while avoiding overfitting. Scheduled training is a soft technique. That is, it starts with a TF scheme and, after the model passes the warm-up stage and the network's weights have gone in the correct update direction, it alters  the original network with a specified scheme. Therefore, the model enjoys stability and the probability of overfitting decreases. In our work, we revisit scheduled training and introduce epoch-scheduled training for forecasting.  
 
\section{The Models}
\label{sec:model}
\subsection{Overview}

In this section, we discuss the proposed models. We denote an input sequence to the model as $(x_1,\cdots,x_T)$ and its output sequence as $(y_1,\cdots,y_{T'})$, where $T$ and $T'$ need not be equal. In the case of predicting $\mathcal{P}$, we have $x_t = [H_{s_t},A_t]$ and $y_{t'} = [\mathcal{P}_{t'}]$. 
 The purpose of the models is to calculate the conditional distribution  $ p(y_1,\cdots,y_{T'}| x_1,\cdots,x_T)$. $ T'$ indicates how far into the future the forecast should go. For example, for hourly-resolution data, if the model is a day-ahead energy flux forecast, then $T'=24$. $T $, on the other hand, is a model parameter and must to be tuned (*1*).\footnote{Recall that we use *$i$* to denote the $i$th tunable parameter.} $T$ can be interpreted as the number of recurrent loops in the structure. Seqtoseq structures contain two independent recurrent networks: encoder and decoder. We use the LSTM network as an encoder, consisting of input $i$, cell state $ s$, output $ o$ and forget $f$ gates (nodes) \cite{gers1999learning}. The standard LSTM equations \cite{lipton2015critical,hochreiter1997long} for time-step $t$ are as follows :
\begin{alignat}{1}
	\label{forget}
&	f_t = \text{sigm}   \left( W_{fx}x_t+ W_{fh}h_{t-1}  \right) \\ \label{input}
&	i_t = \text{sigm} \left(W_{ix}x_t+ W_{ih}h_{t-1} \right)\\
&	o_t = \text{sigm} \left(W_{ox}x_t+ W_{oh}h_{t-1} \right)\\ \label{state}
&	\tilde{s}_t = \text{tanh} \left(W_{\tilde{s}x}x_t+ W_{\tilde{s}h}h_{t-1} \right)\\
&	s_t = f_t \odot s_{t-1} + i_t \odot  \tilde{s}_t\\ \label{hidden}
&	h_t = o_t \odot \text{tanh} (s_t), 	  
\end{alignat}
where $ x_t \in \mathbb{R}^d$, $ h_t \in \mathbb{R}^h$, $ f_t \in \mathbb{R}^h$, $ i_t \in \mathbb{R}^h$, $ o_t \in \mathbb{R}^h$, $ s_t \in \mathbb{R}^h$ and $ \tilde{s}_t \in \mathbb{R}^h$ for all $t=1,..,T$ are the input, hidden state, forget activation, input activation, output activation, cell state and auxiliary cell state vectors, respectively, for the LSTM network. $W_{ij}$ is the weight matrices corresponding to the dimensions of the gate vectors $i$ and $j$. The sigmoid function is given by $\text{sigm}(x)=\frac{1}{1+e^{(-x)}}, \  \forall x \in \mathbb{R}$; its  return value is monotonically increasing in the open interval  $(0,1)$. The notation $\odot$ indicates element-wise (Hadamard) matrix product, which exists only if the matrix dimensions are the same. 

The weight matrices can be initialized randomly with Gaussian distribution. Further, $ h_0$ and $s_0$ are initialized by zero vectors. In energy flux forecasting, $d=2$, and $h$ refers to the hidden vector size, which needs to be tuned (*2*). The biases are omitted in the formulas because one may incorporate them easily through the weight matrices by adding one extra element to each vector mentioned.

A forget gate responds to the question, ``which data should the model keep from the previous cell state vector?'' In \eqref{forget}, the sigmoid function outputs a number between 0 and 1, which indicates the importance of the previous plus current inputs. This value is directly multiplied with the cell state vector of the previous cell state in the first term of \eqref{state}. The input gate \eqref{input} decides which new information should be collected. The second term of \eqref{state} updates this new information. The updated cell state vector is ready to go through the next part of the network. Furthermore, the hidden state is merged into the cell state through \eqref{hidden}.

The purpose of the encoder is to find a representation of the $(x_1,\cdots,x_T)$ sequence as a fixed-sized vector $v$. For the "LSTM+FCL" network, we create a single fully connected layer that receives a fixed-sized vector $v$ as its input and outputs $y_{t'}=[\mathcal{P}_{t'}]$.

For the seqtoseq network, we have another recurrent network as the decoder. The decoder considers vector $ v$ as its input and the last encoder hidden state ($h_T$) as its initial hidden state $h_{t'=0} $ and runs a similar recurrent construction to decode the output sequence. We omit the equations, as they are similar to those above. Furthermore, we let the decoder  update its weight parameters independently. In other words, the encoder and decoder do not share parameters. This doubles the number of parameters to be tuned and multiplies the training time by a constant, but we allow this in the hope of achieving better performance.
The decoder operates a static recurrent structure rather than a dynamic one. That is, the decoder creates an unrolled computational graph of fixed length due to the fixed-sized input vector $v$. In addition, both encoder and decoder can be deep structures. That is, the number of stacked LSTM layers (*3*) for each of them can vary.

\begin{figure}[h]
	\hspace*{-0.75cm}
	\includegraphics[width = 18cm]{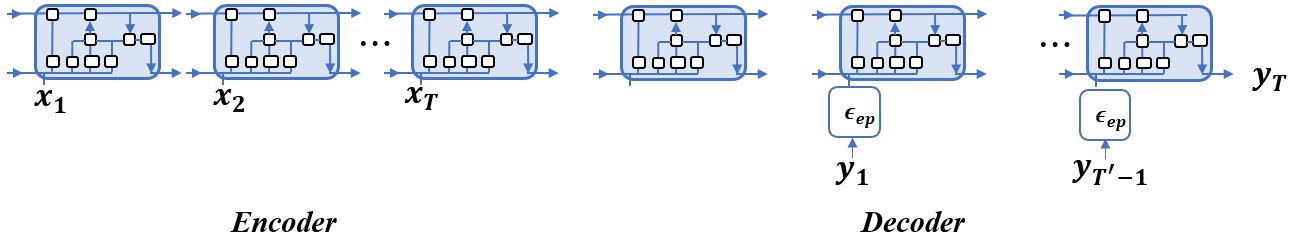}
	\caption{Schematic diagram of the  epoch-scheduled training sequence to sequence network consisting of the encoder and decoder parts.}
	\label{fig:0}
\end{figure}

\begin{figure}[h]
		\centering
	\includegraphics[width = 10cm]{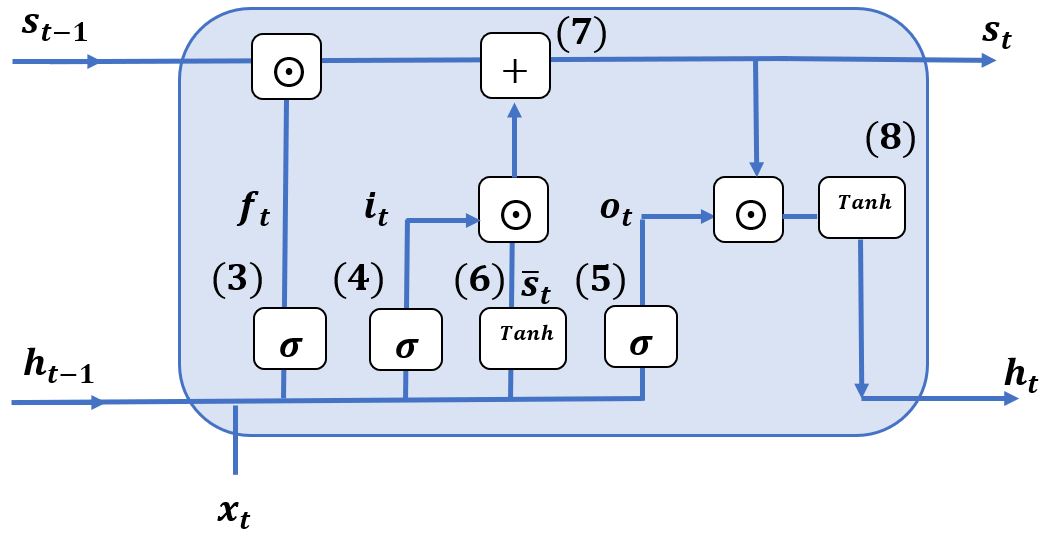}
	\caption{LSTM cell: $\sigma$ and $tanh$ are sigmoid and tangent hyperbolic activation functions, $\odot$ indicates element-wise (Hadamard) matrix product, hidden and state cells propagate through the network.}
	\label{fig:00}
\end{figure}

The seqtoseq model attempts to find the output sequence distribution as follows:
\begin{align}
p(y_1,\cdots,y_T'| x_1,\cdots,x_T;W) & = p(y_1,\cdots, y_T'|v;W) \label{eq:pyx1} \\
& = p(y_1|v;W) \prod_{t'=1}^{T'} p(y_{t'+1}|v,\mathbb{L}\left(y_1,\cdots,y_{t'}\right);W), \label{eq:pyx2}
\end{align}
where $W$ represents all the model weights to be tuned and $\mathbb{L}$ is a binary indicator representing whether the model sees the actual previous measurements or their predictions. The first equality \eqref{eq:pyx1} relies on the encoder and emphasizes once again that the model encodes the whole input $(x_1,\cdots,x_T)$ into the vector $v$ before starting to decode and drops the input thereafter. The concept of multiplying conditional probabilities in the second equality \eqref{eq:pyx2} comes from the recurrent structure of the decoder. In other words, based on the trained weights and the vector $v$, the model tries to forecast the first token (element) of the output sequence. In general, at time step $t'+1$, the model has access to $\mathbb{L}\left(y_1,\cdots,y_{t'}\right)$ of the output sequence. In the TF strategy \cite{williams1989learning,lamb2016professor}, the model has access to the actual true outputs at training time (i.e., $\mathbb{L}\left(y_1,\cdots,y_{t'}\right)=\left(y_1, \cdots,y_{t'}\right)$) and then to the predicted ones (i.e. $\mathbb{L}\left(y_1,\cdots,y_{t'}\right)=\left(\hat{y}_1,\cdots,\hat{y}_{t'}\right)$) at testing time. In the scheduled strategy, however, one may flip a coin \cite{bengio2015scheduled} with probability $\epsilon_t'$ for all $t'=1,\cdots, T'-1$ to decide  whether to use the actual output (with probability $\epsilon_t'$) or its prediction by the model itself (with probability $1- \epsilon_t'$). For instance, the probability of having all true outputs $\left(y_1,\cdots,y_{t'}\right)$ during training is $ \epsilon_1\epsilon_2 \cdots \epsilon_{T'}$. This scheme has been introduced for Machine Translation (MT) tasks, where the number of possible output tokens is as large as the dictionary size. Furthermore, in the MT framework the model faces embedded input and sequence. In contrast, in a forecasting framework, each output token belongs to $\mathbb{R}$. Therefore, using $ \epsilon_{t'}$ for each token $t'$ may result in a combination of true outputs along with predicted ones, which is not particularly useful. Instead, we flip a coin at the beginning of each epoch and stick to the plan for the entire epoch. So, from now on we denote the sequence as $\{\epsilon_{ep}\}$, where $ep$ is the training epoch number. Intuitively, the sequence  $\{\epsilon_{ep}\}$ should be decreasing, which encourages the model to use the predicted output towards the end of the training. \cite{bengio2015scheduled} introduces Linear, Exponential and Inverse Sigmoid Decay sequences. We modify the Inverse Sigmoid Decay to use for epoch-scheduled training as follows:
$$ \epsilon_{ep}= \frac{k}{k+e^{(ep/k)}},$$	
where $k$ is a parameter to tune (*4*). Increasing $k$ increase the probability of receiving true values. For example, for a training scheme of size 20 epochs, $k=20$ means that we receive the true outputs with at least 0.86 probability. Hence, in the tuning process, we compare the magnitude of $k$ with the number of epochs.\footnote{However, we manually force the model to use the predicted values for the last two epochs.} Figure \ref{fig:0} illustrates a schematic of the epoch-scheduled training sequence to sequence network. An expanded view of a single LSTM cell structure is presented in Figure \ref{fig:00}, with details and equations (\ref{forget})-(\ref{hidden}) marked in the figure. Hidden and state cells propagate through the encoder and decoder networks. In Figure \ref{fig:0}, $\epsilon_{ep}$ expresses the probability that a given decoder cell sees the actual outputs, where $ep$ index iterates over all epochs.

\subsection{Loss Function \& Optimizer}

We use Mean Squared Error (MSE) plus a 2-norm regularizer as the loss function for the whole network:
\begin{equation} \label{mse}
 \text{MSE}=\frac{1}{n} \sum_{i=1}^{n} \left(y^i -\hat{y}^i \right)^2,
\end{equation}
where $n$ is the size of the measurements and $\hat{y}^i$ is a vector of size $T'$ representing the model's prediction of $y^i$, which is the actual power outputs. The data, however, is chopped up and fed via mini batches, the size of which is tunable (*5*) as a power of 2, which is a common practice in optimization algorithms. Regardless of the algorithm used, MSE is clearly convex with respect to $\hat{y}^i$. But we do not have direct control over $\hat{y}^i$. Instead, $\hat{y}^i = N(x^i;W) $ where $N$ is the seqtoseq network structure and $x^i$ is the $i$-th network input. The loss function is generally non-convex with respect to $W$,  which justifies our use of a 2-norm regularizer \cite{nie2010efficient,liu2009multi}. Hence, a term  $\lambda \|w \|^2_2$ is added to the loss function, where $w$ is a vector that stacks all the trainable parameters within the weight matrices and $\lambda$ (*6*) is a regularization parameter to be tuned. Therefore, the loss function equals
$$ f_w(x)= \frac{1}{n} \sum_{i=1}^{n} \left(y^i -N(x^i;w) \right)^2 + \lambda \|w \|^2_2.$$        
Through backpropagation, we calculate the gradient of the loss function and update the weights via a first-order optimization algorithm. Stochastic Gradient Descent (SGD) directly updates $w$ in the direction of the negative of the mini-batches' gradient in an iterative manner. That is, if $b_i$ is the $i$th mini-batch, then we have
$$ w_{i+1}:= w_i - \alpha_i \nabla_w \bar{f}_w(x), $$    
where 
\[\bar{f}_w(x)=\frac{1}{|b_i|} \sum_{b_{i}} \left(y^i -N(x^i;w) \right)^2 + \lambda \|w \|^2_2\]
 and $\alpha_i$ is the learning rate (*7*) at step $i$, which needs to be tuned. There exist several modifications of SGD, such as RMSProp.  Adaptive moment estimation (Adam) \cite{kingma2014adam} has been successfully implemented as another first-order method. In Adam, the model's weight is updated as follows:

\begin{equation}
w_{i+1}= w_i - \frac{\alpha_i }{\sqrt{\hat{v}_i}+\epsilon} \hat{m}_i,
\end{equation}

where
\begin{alignat}{2}
&\hat{m}_i=\frac{m_i}{1-\beta_1^i}\\
&\hat{v}_i=\frac{v_i}{1-\beta_2^i}\\
&m_i:= \beta_1m_{i-1}+ (1-\beta_1)\nabla_w\bar{f}_w(x)\\
&v_i:= \beta_2m_{i-1}+ (1-\beta_2)\nabla_w^2\bar{f}_w(x),
\end{alignat}
$\beta_1$ and $\beta_2$ are momentum-like parameters, and $\epsilon$ serves to reduce numerical issues. Adam aims to update each element of $w$ with respect to its size (adaptively); it stores only an exponentially decaying average of past squared gradients. The authors of \cite{kingma2014adam} propose $\beta_1=0.9$, $\beta_2=0.99$, and $\epsilon=1\mathrm{e}{-8}$ to be efficient. In some cases, such as machine translation, in which we are dealing with RNN structures, similar to our forecasting context, Adam suffers from using exponential averaging over the past squared gradients \cite{reddi2018convergence2}. AMSGrad has been proposed to tackle this issue by simply using the maximum of the past squared gradients. In other words, we update $\hat{v}_i= \max\{\hat{v}_{i-1},v_i\}$.
Below, we first compare the efficiency of the SGD, RMSProp, Adam, and AMSGrad algorithms, and then optimize the energy flux forecasting model mostly with Adam and AMSGrad.

\subsection{Data}
Satellite altimeters and buoy measurements are the two most common sources of data for wave feature forecasting \cite{cuadra2016computational}. In this study, we primarily use buoy measurements from the National Oceanic and Atmospheric Administration (NOAA). Each buoy provides measurements of significant wave height, wind speed, wind direction, average wave period ($A$), sea level pressure, gust speed, air temperature, and sea surface temperature, at resolutions of  10 minutes to 1 hour.\\
The National Data Buoy Center (NDBC) maintains three major data sets, consisting of moored buoys, drifting buoys, and Coastal-Marine Automated Network (C-MAN) stations, which are located alongside U.S.\ coastal structures. Along with other air and water features, they monitor wave energy spectra, from which $H_s$ and $A$ can be obtained via equation \eqref{momentum}. Each site has an  identifier (ID). IDs are in the form of five digits, except for C-MAN stations, which have alphanumeric IDs. The first two digits are assigned to a continental region, and the last three indicate a specific location. For instance 41001, 41002 and 41004 are Atlantic Ocean sites near the southeastern U.S. The sites  typically have hourly resolution, which produces 8760 $H_s$ and $A$ data points per year. The data sets use 99.00 to indicate missing measurements. Some sites, however, aimed to report 10-minute-resolution data in 2017. For instance, among the above-mentioned stations, 41002 continues to provide  hourly data, while 41001 and 41004 try to provide 10-minute-resolution data, but for the most part, their data is still hourly, with the non-hourly values filled by 99.00. Quite a few refined stations have been collected for this study. We consider a buoy to be a ``refined station'' when it is active and has a year with at least 1000 meaningful data points (approximately 11.4\%).
We calculate energy flux $\mathcal{P}$ via equation \eqref{power} for the datasets that remain after this process.
Table \ref{tab1} displays the  buoys investigated in this research, along with some information about them. The locations are chosen from nearshore South Pacific Ocean, the Gulf of Mexico, and the North Atlantic Ocean, where the water depths range from approximately 100 m to 5 km. We divided the data into 3 parts: 60\% for training, 20\% for hyperparameter tuning and 20\% for testing unless we are comparing with alternative approaches and they used another division. The model has no access to the testing data in any manner. Table \ref{tab1} indicates the training and testing data sizes as well. The validation (tuning) is similar to testing hence we did not show that.

\begin{table}[h]
	\caption{Information of selected refined buoys}
	\label{tab1}
		\setlength\tabcolsep{3pt}
		\hspace*{-0.25cm}
	\begin{tabular}{cccccccc}
Usage                 & Station & Lon     & Lat    & Water Depth & Years                          & Training Points &  Test Points \\		       
		              & ID      & (W)     & (N)    & (m)         & Interval                       & \#              & \#                        \\\hline
Forecasting:          & 41043   & 64.830  & 21.124 & 5271        & {[}2007,2017{]}                & 52911           & 17904                     \\
		              & 41040   & 53.045  & 14.554 & 5112        & {[}2005,2017{]}                & 54794           & 18617                     \\
		              & 42056   & 84.938  & 19.918 & 4526        & {[}2005,2017{]}                & 58206           & 19548                     \\
		              & 32012   & 85.078  & 19.425 & 4534        & {[}2007,2017{]}                & 49280           & 16428                     \\
		              & 41049   & 62.938  & 27.490 & 5459        & {[}2009,2017{]}                & 44816           & 14996                     \\
		              & 41060   & 51.017  & 14.824 & 5021        & {[}2012,2017{]}                & 24817           & 8272                      \\
		              & 41001   & 72.617  & 34.625 & 4453        & {[}1976,2017{]}\textbackslash\{1979,2008\}$^1$   & 134018          & 44791                     \\
		              & 51002   & 157.696 & 17.037 & 4934        & {[}1984,2017{]}\textbackslash\{2013\}$^1$        & 132560          & 44507                     \\
		    		  & 46086   & 118.036 & 32.507 & 1838        & {[}2003,2017{]}                & 69044           & 23233                     \\
		   		  	  & 46013   & 123.307 & 38.238 & 122.5       & {[}1981,2017{]}                & 164022          & 54980                     \\
		   			  & 41048   & 69.590  & 31.860 & 5340        & {[}2007,2017{]}                & 49820           & 16705                     \\
		   			  & 46084   & 136.102 & 56.622 & 1158        & {[}2002,2017{]}\textbackslash\{2009\}$^1$        & 64186           & 21606                     \\             
		   			  & 46083   & 137.997 & 58.300 & 136         & {[}2001,2017{]}\textbackslash\{2014\}$^1$        & 58440           & 18273                     \\
Reconstruction:       & 46042   & 122.398 & 36.785 & 1645.9      & {[}2009,2010{]}                & 4687            & 4576                     \\
			          & 46025   & 119.049 & 33.761 & 888         & {[}2009,2010{]}                & 4687            & 4576                     \\
		              & 46069   & 120.229 & 33.663 & 986         & {[}2009,2010{]}                & 4687            & 4576                     \\
Feature Selection:    & 44007   & 70.141  & 43.525 & 26.5        & {[}2018{]}                     & 2737            & 685                     \\
				      & 44008   & 69.248  & 40.504 & 74.7        & {[}2018{]}                     & 2737            & 685                     \\
				      & 44009   & 74.702  & 38.457 & 30          & {[}2018{]}                     & 2737            & 685                     \\
				      & 44013   & 70.651  & 42.346 & 64          & {[}2018{]}                     & 2737            & 685                     \\	
				      & 44014   & 74.840  & 36.606 & 47          & {[}2018{]}                     & 2737            & 685                \\	
				      & 44017   & 72.049  & 40.693 & 48          & {[}2018{]}                     & 2737            & 685 \\					      			      				      			      
				      				      
\hline
	\end{tabular}

{  $1$: These years are excluded based on unsatisfactory amount of available data}
\end{table}
We argue that some of the studies in the literature are insufficient from a data point of view in two broad senses. First, for papers that use real buoy measurements, there is usually extensive data manipulation and preprocessing. Sometimes the proposed models see only subsamples of measurements in order to produce smaller forecasting metric errors.    
For example, \cite{kumar2018ocean,kumar2018ocean2} choose subsample of 50 daily points out of 8 months of data (Jan 1--Aug 30, 2015) and report Root Mean Squared Error (RMSE) as an error measurement metric. Not only is this metric scale-dependent and can be changed by shrinking the data size, but this selection of points can break the temporal notion of the data as well. Therefore it has a direct effect on the nonlinear functional relation that we aim to predict. Moreover, \cite{hatalis2014multi} produce the data using the power matrix method. Hence, the accuracy of the method heavily depends on the accuracy of the data-producing procedure. In our study, after selecting a buoy, we do not conduct \textit{any} data manipulation or preprocessing, other than cleaning not-a-number (NaN) values from the NOAA data. Otherwise, the input to the model is the raw data collected from NOAA. (See \url{https://www.ndbc.noaa.gov/wave.shtml} to see how spectral wave data are derived from buoy motion measurements.)

\subsection{Error Metrics}
We use the following forecasting Error Metrics:
\begin{itemize}
	\item Root Mean Square Error (RMSE) is one of the most commonly used regression error metrics. It equals the square root of the MSE, given in \eqref{mse}, and it provides an useful measure of forecasting quality. The closer the RMSE gets to 0, the better the fit the prediction gives. In general, it is difficult to choose an appropriate threshold such that the prediction is deemed ``accurate'' if the RMSE is less than that threshold, because because the RMSE is scale dependent and not robust to outliers.
	\item HUBER loss, given two sets of points $ y$ and $\hat{y}$, is defined as 
\begin{equation}
	H_{\delta}= 	\frac{1}{n} \sum_{i=1}^{n} \mathcal{L}_\delta \left(y^i ,\hat{y}^i \right),
\end{equation}

	where
	\begin{alignat}{1}
\mathcal{L}_\delta \left(y^i ,\hat{y}^i \right)= \left\{
\begin{array}{ll}
\frac{1}{2}(y^i -\hat{y}^i)^2  & \mbox{for } |y^i -\hat{y}^i | \leq \delta \\
\delta | y^i -\hat{y}^i|- \frac{1}{2}\delta^2 & \mbox{otherwise.} 
\end{array}
\right.
	\end{alignat}
	Generally, $\delta=1$ is an acceptable threshold. HUBER loss is scale dependent but is less sensitive to outliers than RMSE is. 
	\item Pearson Correlation Coefficient (CC) is defined as:
\begin{equation}
	\rho_{y,\hat{y}}= \frac{\text{cov}(y,\hat{y})}   {\sigma_y\sigma_{\hat{y}}},
\end{equation}

	 where $\text{cov}(\cdot)$ is covariance.
	 We have $-1\leq CC \leq 1$. Although CC captures linear similarities of its inputs very well, several characteristics of nonlinear relations are ignored.
	\item Mean Arctangent Absolute Percentage Error (MAAPE), given two sets of points $ y$ and $\hat{y}$, is defined as:
\begin{equation}
	\frac{1}{n} \sum_{i=1}^{n} arct \left(|\frac{y^i -\hat{y}^i}{y^i}| \right).
\end{equation}

	MAAPE is scale-independent and, unlike MAPE, overcomes problematic cases as $y^i$ goes to zero for all $i=1,\cdots,n$. 
\end{itemize}

\subsection{Hyperparameter Tuning}
In previous sections we identified tunable parameters ($^*i^*$), $i=1,2,...,7$. Here, we use Spearmint Bayesian optimization \cite{snoek2012practical} as a tuning tool. The algorithm is capable of handling integer parameters, as in our case \cite{swersky2013multi}. The algorithm treats the seqtoseq forecasting model as a random black box function and places a Gaussian Process (GP)  prior over it. After collecting function evaluations, it extracts posterior information based on the Expected Improvement (EI) observed. Hence, GP-EI-OptChooser with Monte Carlo approximation is selected, which tells Spearmint which of the candidates to execute next. The values of each parameter that we consider are listed in Table~\ref{tab:0}.

\begin{table}[h]
	\label{tab:0}
	\centering
	\caption{Possible values for parameters of the sequence-to-sequence model.}
	\begin{tabular}{ll}
		hidden size        & \{32,64\}                    \\ 
		time step         &  \{10,20,30,40,50,60\}                         \\ 
		batch size  & \{16,32,64,128,256\}               \\ 
		learning rate $\alpha_i$       & \{0.001,0.002,0.003,0.004,0.005,0.006,0.007,0.008,0.009,0.010\}                           \\ 
		\# stacked layers     & \{1,2,4\}              \\ 
		2-norm regularizer $\lambda$              & \{0.001,0.002,0.003,0.004,0.005,0.006,0.007,0.008,0.009,0.010\}         \\ 
		$k$ scheduled training       & \{0.1,0.2,0.3,0.4,0.5,0.6,0.7,0.8,0.9,1.0\} $\times$ epochs                       \\ 
	\end{tabular}
\end{table} 

\begin{figure}[htbp]
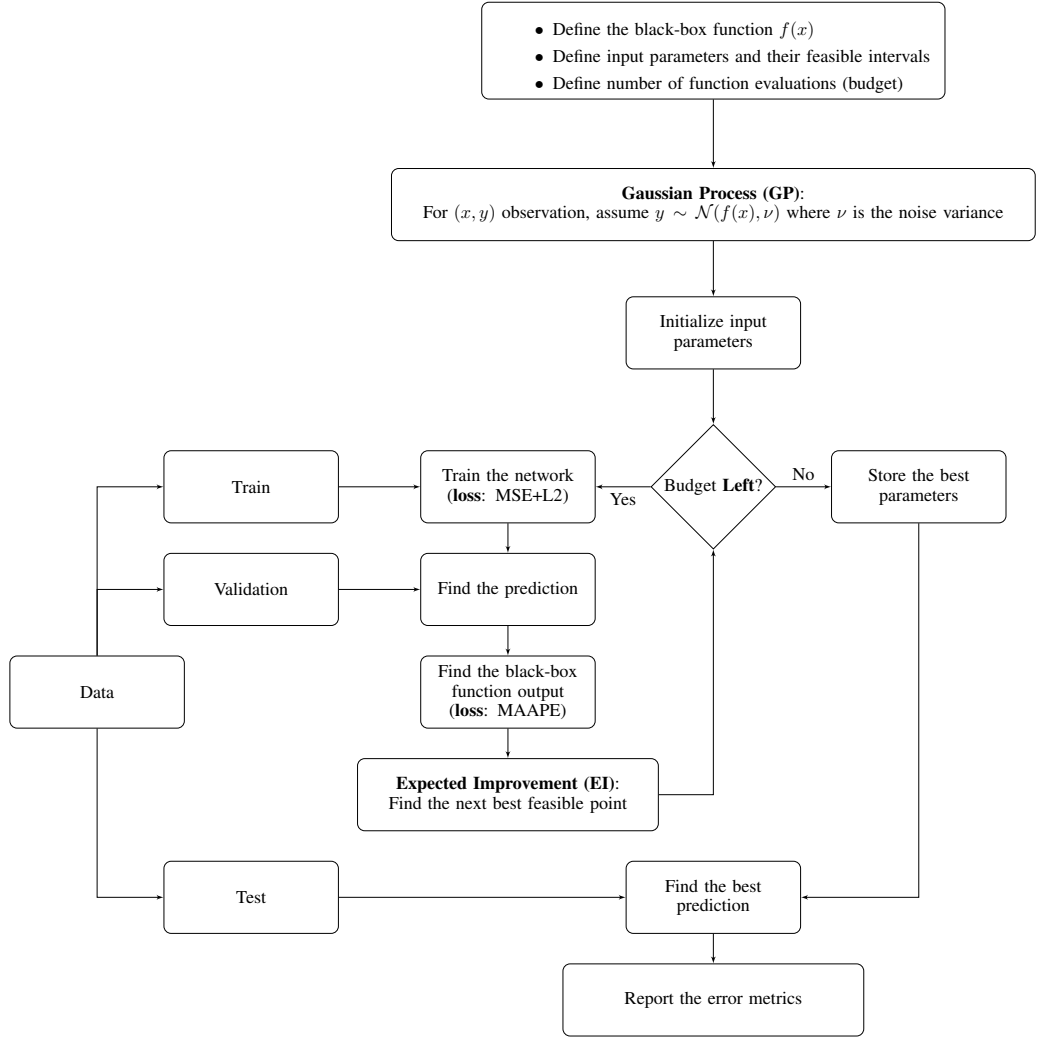

		\centering
	\includestandalone[width = 14cm]{hyperParameter}
	\caption{Hyperparameter tuning process with Spearmint Bayesian optimization (GP-EI-OptChooser) on black-box (seqtoseq) function}
	\label{Hyper}
\end{figure}

Spearmint only has access to the validation dataset and the model that has already been trained on the training dataset. Any hyperparameter tuning algorithm such as Spearmint can have its own objective function. That is, Spearmint's objective function during validation can be separate from the seqtoseq model's during training (which, recall, is MSE plus a 2-norm regularizer). That is, the model enjoys (MSE plus 2-norm regularization) convergence properties in training while enjoys using MAAPE scale independency and upper bound ($\pi/2 $) during the tuning process. Figure \ref{Hyper} illustrates the process of using Spearmint for hyperparameter tuning. We call the number of function evaluations as the ``algorithm budget.''

\section{Experimental Results}
\label{sec:exp}
In this section, we first report on the model's performance utilizing several optimization algorithms and compare them to those of standard Neural Networks on buoy measurements. Next, we demonstrate the performance of our model for long-term forecasting (48 steps ahead) of $H_s$. Finally, we tackle the problem of constructing  wave characteristics such as $H_s$ for unknown locations based on  information from known locations. We wrote the code for our method and its alternatives in Python 3.6 using the TensorFlow package (running on a GeForce 1050 GTX GPU), except the code for hyperparameter tuning via Spearmint, whose most stable version for our framework is written in Python 2.7.  

\subsection{Comparison of Optimization Algorithms}
\begin{sidewaysfigure}[htbp]
	\hspace*{-3.75cm}
	\scalebox{0.62}{\input{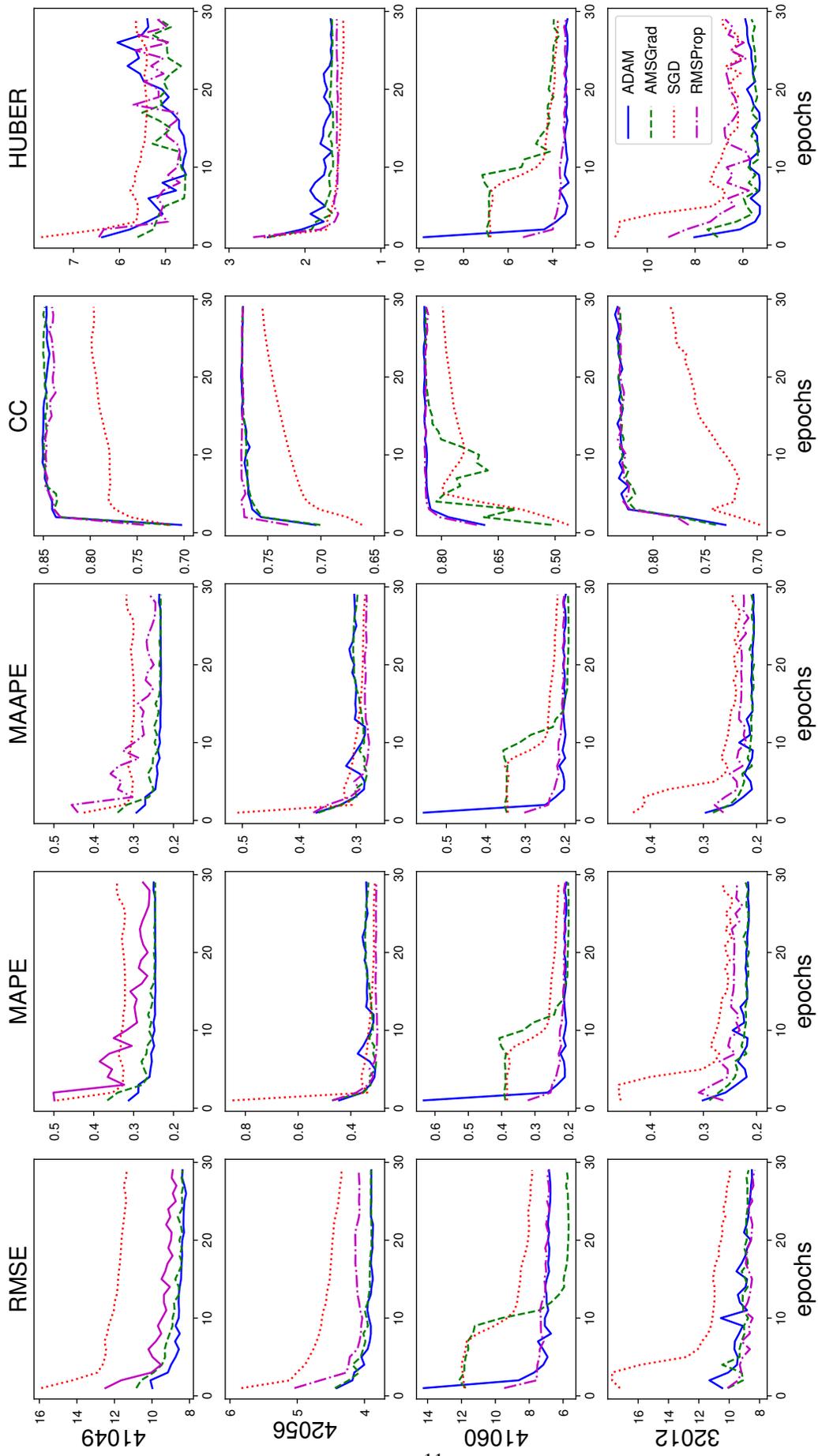}}
	\caption{Test error comparison for four measurement stations. Learning rate: $ \alpha=0.001$ with decay=$0.9$ for all,  Adam \& AMSGrad: $\beta_1=0.9$ and $\beta_2=0.99$ and $\epsilon=1\mathrm{e}{-8}$; RMSProp: momentum=$0.0$,$\epsilon=1\mathrm{e}{-10}$.}
	\label{fig:1}
\end{sidewaysfigure}

In this section, we aim to compare the effectiveness of the SGD, RMSProp, Adam, and AMSGrad optimization schemes based on test error for four of the refined buoys. Figure \ref{fig:1} displays the algorithm test errors versus the number of training epochs, considering all five error metrics introduced earlier. For the cases of RMSE and HUBER losses, the values on the y-axis do not carry a clear meaning. In other words, we only hope to reach the smallest possible values for these two loss functions, but the actual values do not tell us much. One can observe that, even at the end of the training epochs, RMSE and HUBER losses are not negligible. Note that the error we experience during the calculation of $\mathcal{P}$ is of the third power of the error we encounter forecasting $H_s$. A similar difference has been pointed out by \cite{cornejo2018bayesian}. Moreover, RMSE and HUBER are both scale-dependent, and the test sets include years of data, which may result in errors piling up over time. In contrast, MAPE and MAAPE are scale independent, and therefore their values have more intuitive meaning. In addition, their behavior is similar to each other, considering that MAAPE values are consistently lower than those of  MAPE. Therefore, we mainly focus on MAPE and MAAPE for subsequent experiments. 
SGD and, in a few cases, AMSGrad require more epochs to reach their best test accuracy, while Adam usually reaches its peak faster. RMSProp and Adam perform similarly, with Adam experiencing a marginal lead. AMSGrad and Adam exhibit excellent performance in terms of RMSE error. We used a learning rate of $\alpha=0.001$ with a decay of $0.9$ for all the algorithms. This is to make the algorithms more stable; otherwise, having a fixed learning rate would cause most algorithms, especially SGD, to encounter drastic changes towards the end of the optimization process. HUBER losses are the most challenging loss functions to deal with. One can see sizable fluctuations, even near  the end of the training, for stations 41049 and 32012. SGD shows inferiority in terms of test error; however, based on its simple update, it enjoys the best results in terms of time needed for training the algorithms. SGD is especially sensitive with respect to its parameters. Adam, however, expresses stability and is among the first algorithms to converge to its best result (usually within its first 10 training epochs). Although RMSProp performs particularly well for station 42056, there are cases where the algorithm cannot reach the desired test error. Based on these results, from now on we use either Adam or AMSGrad as the optimizer.  

\subsection{Networks Comparison}
\label{sec:network_comparison}

In this section, we conduct an experiment to evaluate the performance of different neural network structures on the performance of the algorithm. We compare Single (Multi) layered RNN, Single (Multi) layered LSTM and epoch-scheduled seqtoseq model with Adam and AMSGrad optimization algorithms. RNN and LSTM networks contain a fully connected last layer. We consider seqtoseq model with both Adam and AMSGrad optimization algorithms.

Tables \ref{tab2} and \ref{tab3} demonstrate the performance of these neural network structures for 5- and 10-step-ahead forecasting of energy flux over data gathered from 5 refined buoys (collected from Table \ref{tab1}), using MAPE and MAAPE as the error metrics.

To evaluate the networks fairly, during hyperparameter tuning we used the same budget in Spearmint for each network. We set this budget equal to 100 function evaluations. One may argue that this experimental design might favor the alternate networks. Because they have many fewer hyperparameters to tune but for seqtoseq framework Spearmint deals with a more complex function to extract information from, having the same number of function evaluations. 

Multi-layered networks consist of four layers. The first three layers are recurrent of structure [time steps, time steps, hidden size] with REctified Linear Unit (RELU) activation functions and the last layer is a fully connected one attaching the previous output to a layer of the last node. Training process has done with the budget of at most 15 epochs. In other words, if a structure reaches its best performance anywhere before 15th epoch, that would be collected. The best parameters for epoch-scheduled seqtoseq are $[64,10,16,0.001,1,0.001,0.4\times \text{epochs}]$ in same order as shown in table \ref{tab:0}.

From Table~\ref{tab2}, one can see that seqtoseq networks result in smaller MAPE and MAAPE values, but single-layered LSTM plus a fully connected last layer also performs very well. We believe a fully connected last layer assists LSTM and simple RNN networks in capturing all learned features and translating them into superior forecasting. In addition, Multi layered structures are not superior to single layered ones. This happens in the Spearmint tuning process as well. Spearmint mostly prefers single or double layer structures for energy flux forecasting of any structures. Hence, when we force the stacked layers to be four, we can see a tiny reduction in the performance of the models. Moreover, Adam and AMSGrad perform closely to each other, and one cannot claim the superiority of any.

\begin{table}[h]

		\caption{5-step-ahead energy flux forecasting, recurrent time step =$10$, hidden size=$64$, stacked recurrent layers: $1$ or $4$, batch size=$16$. }
	\label{tab2}
		\setlength\tabcolsep{1pt}
		\hspace*{-0.5cm}
	\begin{tabular}{ccccccccccccccccccc} 		\hline  
		     
		Station ID &  & \multicolumn{2}{c}{SL-RNN+FCL} &  & \multicolumn{2}{c}{SL-LSTM+FCL} &  & \multicolumn{2}{c}{ML-RNN+FCL} &  & \multicolumn{2}{c}{ML-LSTM+FCL} &  & \multicolumn{2}{c}{Seqtoseq(Adam) } &  & \multicolumn{2}{c}{Seqtoseq (AMSGrad) } \\ \cline{1-1} \cline{3-4} \cline{6-7} \cline{9-10} \cline{12-13} \cline{15-16} \cline{18-19} 
		&  & MAPE       & MAAPE       &  & MAPE        & MAAPE       &  & MAPE       & MAAPE       &  & MAPE        & MAAPE       &  & MAPE         & MAAPE        &  & MAPE         & MAAPE        \\ \cline{3-4} \cline{6-7} \cline{9-10} \cline{12-13} \cline{15-16} \cline{18-19} 
		46013      &  & 0.296       & 0.259       &  & 0.295        & 0.258       &  & 0.303       & 0.266       &  & 0.301        & 0.265       &  & {\bf0.269}         & {\bf0.248}        &  & 0.297         & 0.271        \\
		41048      &  & 0.336       & 0.279       &  & 0.308        & 0.259       &  & 0.334       & 0.278       &  & 0.354        & 0.289       &  & {\bf0.217}         & {\bf0.205}        &  & 0.221         & 0.211        \\
		46084      &  & 0.395       & 0.306       &  & 0.392        & 0.304       &  & 0.397       & 0.307       &  & 0.399        & 0.308       &  & 0.311         & 0.279        &  & {\bf0.281}         & {\bf0.258}        \\
		46083      &  & 0.389       & 0.310       &  & 0.402        & 0.318       &  & 0.397       & 0.314       &  & 0.464        & 0.356       &  & 0.291         & 0.266        &  & {\bf0.271}         & {\bf0.253}        \\
		41060      &  & 0.215       & 0.193       &  & 0.221        & 0.197       &  & 0.209       & 0.189       &  & 0.201        & 0.180       &  & 0.174         & 0.169        &  & {\bf0.163}         & {\bf0.159}\\
		\hline       
	\end{tabular}
\end{table}

In Table~\ref{tab3}, we can observe that when we move from 5- to 10-step-ahead forecasting, epoch-scheduled seqtoseq fits the actual outputs even more accurately. For instance, at station 41048, seqtoseq provides a $(0.308-0.217)/0.217=  41.9\%$ improvement over the best alternative method in terms of MAPE for 5-step-ahead forecasting, whereas for 10-step-ahead the improvement is $(0.459-0.299)/0.299=53.5\%$. Similarly, for MAAPE, the 5- and 10-step improvements are $(0.259-0.205)/0.205)=26.3\%$ and $(0.333-0.278)/0.278=33.2\%$, respectively.  These extra improvements may be a direct result of scheduled training managing deeper forecasting horizons. 

\begin{table}[h]
	\caption{ 10-step-ahead energy flux forecasting, recurrent time step =$10$, hidden size=$64$, stacked recurrent layers: $1$ or $4$, batch size=$16$.}
	\label{tab3}
	\setlength\tabcolsep{1pt}
		\hspace*{-0.5cm}
	\begin{tabular}{ccccccccccccccccccc}		\hline       
		Station ID &  & \multicolumn{2}{c}{SL-RNN+FCL} &  & \multicolumn{2}{c}{SL-LSTM+FCL} &  & \multicolumn{2}{c}{ML-RNN+FCL} &  & \multicolumn{2}{c}{ML-LSTM+FCL} &  & \multicolumn{2}{c}{Seqtoseq(Adam) } &  & \multicolumn{2}{c}{Seqtoseq (AMSGrad) } \\ \cline{1-1} \cline{3-4} \cline{6-7} \cline{9-10} \cline{12-13} \cline{15-16} \cline{18-19} 
		&  & MAPE       & MAAPE       &  & MAPE        & MAAPE       &  & MAPE       & MAAPE       &  & MAPE        & MAAPE       &  & MAPE         & MAAPE        &  & MAPE         & MAAPE        \\ \cline{3-4} \cline{6-7} \cline{9-10} \cline{12-13} \cline{15-16} \cline{18-19} 
		46013      &  & 0.487          & 0.360         &  & 0.459          & 0.347          &  & 0.473             & 0.359             &  & 0.473              & 0.359             &  & {\bf 0.376}                 &{\bf 0.328} &  & 0.402 & 0.345 \\
		41048    &  & 0.470          & 0.336         &  & 0.459          & 0.333          &  & 0.707             & 0.470             &  & 0.691              & 0.450             &  & 0.436                 & 0.345      &&  {\bf0.299}   &  {\bf0.278}   \\
		46084      &  & 0.734          & 0.434         &  & 0.742          & 0.437          &  & 0.769             & 0.454             &  & 0.727              & 0.432             &  & 0.494                 & 0.377        &&  {\bf0.393}   & {\bf0.341}  \\
		46083      &  & 0.592          & 0.399         &  & 0.569          & 0.393          &  & 0.624             & 0.410             &  & 0.602              & 0.406             &  & 0.425                 & 0.360          &&  {\bf0.388}  & {\bf0.341} \\
		41060      &  & 0.353          & 0.282         &  & 0.375          & 0.295          &  & 0.374             & 0.298             &  & 0.428              & 0.331             &  & {\bf0.207}                 & {\bf0.199}   &&  0.209  & 0.201  \\
		\hline       
	\end{tabular}

\end{table}

\subsection{Long-Term Forecasting}

Next, we investigate 2-day-ahead forecasting of $H_s$. As noted by \cite{dixit2016prediction}, 2 days can be considered an acceptable range for long-term forecasting of $H_s$. Table \ref{tab4} presents 48-step-ahead test error comparison over 5 refined buoys for $H_s$ forecasting using the seqtoseq model.  

\begin{table}[]
		\centering
		\caption{ 48 steps ahead  $H_s$ forecasting, recurrent time step =$10$, hidden size=$64$, stacked recurrent layers: $2$, batch size=$16$.}
	\label{tab4}
		\setlength\tabcolsep{3pt}
\begin{tabular}{lcccccccccccccccccc}		\hline       
	Station ID &  & \multicolumn{7}{c}{Seqtoseq (Adam)}    &  & \multicolumn{8}{c}{Seqtoseq (AMSGrad)}    &  \\ \cline{1-1} \cline{3-9} \cline{11-18}
	&  & RMSE (m)  &  & MAPE  &  & MAAPE &  & HUBER &  & RMSE (m) &  & MAPE  &  & MAAPE &  & HUBER &  &  \\ \cline{3-3} \cline{5-5} \cline{7-7} \cline{9-9} \cline{11-11} \cline{13-13} \cline{15-15} \cline{17-17}
	41043      &  & {\bf0.593} &  & 0.249 &  & 0.235 &  & {\bf0.147} &  & 0.598 &  & {\bf0.247} &  & {\bf0.233} &  & 0.150 &  &  \\
	41040      &  & 0.431 &  & {\bf0.182} &  & 0.177 &  & {\bf0.089} &  & 0.431 &  & 0.183 &  & {\bf0.176} &  & 0.090 &  &  \\
	41001      &  & 1.197 &  & 0.459 &  & 0.343 &  & 0.457 &  & 1.197 &  & 0.459 &  & 0.343 &  & 0.457 &  &  \\
	51002      &  & 0.513 &  & 0.186 &  & 0.180 &  & 0.126 &  & {\bf0.511} &  & 0.186 &  & {\bf0.179} &  & {\bf0.125} &  &  \\
	46086      &  & {\bf0.658} &  & {\bf0.238} &  & {\bf0.226} &  & {\bf0.177} &  & 0.666 &  & 0.240 &  & 0.230 &  & 0.181 &  & \\		\hline       
\end{tabular}
\end{table}

Note the large gap between RMSE and HUBER losses that the model experiences when it predicts energy flux compared to significant wave height. The reason is the difference in the value range of output power and significant wave height. $\mathcal{P}$ is on the order of 10 kW/m, while $H_s$ is usually on the order of 1 m. Hence, we focus on MAAPE. The results indicate that the model is robust for long-term prediction. The values presented in Table \ref{tab4} are comparable to those for 5- and 10-step-ahead predictions for $\mathcal{P}$ in Section~\ref{sec:network_comparison}. Further, by definition, the HUBER metric should behave robustly with regard to outliers. The results support this fact and the model enjoys minimal HUBER loss values.    

From the table, it is clear that Adam and AMSGrad produce very similar errors. Therefore, we conclude that one will not miss useful details by considering only the Adam optimization algorithm. Hence, in the last section, we optimize the models using Adam.

\section{Reconstruction of Significant Wave Height}
\label{sec:recon}
\begin{figure}[htbp]
		\centering
	\includegraphics[width = 9cm]{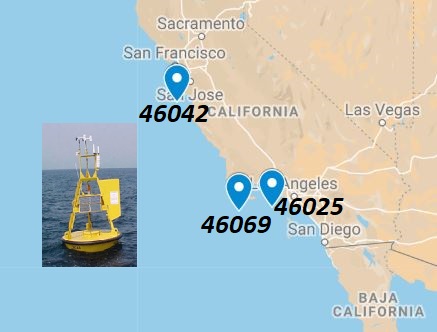}
	\caption{Three adjacent buoys of 46069, 46042 and 46025}
	\label{fig:buoys}
\end{figure}

Our framework, on its surface, is not designed to reconstruct missing measurements of a station based on information from other stations; rather, it uses historical data at the same station. Therefore, to address this goal, we modify the input of the neural networks in such a way that the models will no longer require historical data. Figure~\ref{fig:buoys} illustrates three buoys---46069, 46025 and 46042---that are close to one another. For more details on buoys you can see the Table \ref{tab1}. Similar to existing methods \cite{cornejo2016significant,cornejo2018bayesian}, we aim to reconstruct the significant wave height of  buoy 46069 at different time steps, treating them as missing data. Networks have access to all information from the two adjacent buoys, 46025 and 46042, as their input, as well as  buoy 46069's SWH as their labels in the training process, which consists of data from the entire year 2009. Then, the networks predict the entire 2010 year SWH of buoy 46069, based on the inputs from the adjacent buoys. In other words, at time step $t$, each model is allowed to see the first $t$ measurements of $H_s$ at the two nearby buoys.

Table \ref{tab6} compares the performance of our seqtoseq and SL LSTM networks with that of benchmark methods from the literature. The methods are \cite{cornejo2016significant,cornejo2018bayesian}: 1) All-featured Extreme Learning Machine, (ELM) 2) All-featured Support Vector Regression (SVR), 3) Grouping Genetic Algorithm Extreme Learning Machine with final prediction with ELM (GGA-ELM-ELM), 4) Grouping Genetic Algorithm Extreme Learning Machine with final prediction with SVR (GGA-ELM-SVR), 5) Bayesian optimization Grouping Genetic Algorithm Extreme Learning Machine with final prediction of ELM (BO-GGA-ELM-ELM) and 6) Bayesian optimization Grouping Genetic Algorithm Extreme Learning Machine with final prediction of SVR (BO-GGA-ELM-SVR). The table reports the RMSE, mean absolute error (MAE), and $r^2$ (which is CC$^2$) as error metrics; these are the metrics used in the benchmark papers.

Note that \cite{cornejo2016significant,cornejo2018bayesian} uses buoys 46025, 46042, and 46069, and claims that the NOAA dataset has no missing measurements for these buoys in the years 2009--10. However, we found this to be untrue. First, there are significant gaps in the 46069 station data, which result in fewer than 5000 data points instead of the 8760 hourly points that should be present in a given year. Second, like the other refined stations, roughly 1--2\% of the $H_s$ values contain 99.00, which indicates missing data. Therefore, we carefully cleaned the missing data points from all three stations in such a way as to preserve the position of each time step. That is, if only there is a missing point in a station measurement, we exclude that time step from all stations. The resulting data set has 9263 data points for each buoy, where 4687 points are for training (2009) and 4576 are for testing (2010). 

\begin{table}[h]
		\caption{ Comparison of different methods for reconstructing $H_s$ of  station 46069, epochs $\leq 50$.}
			\label{tab6}
			\centering
	\begin{tabular}{lllllll}		\hline       
		Methods      &  & RMSE (m)  &  & MAE (m)   &  & $r^2$  \\ \cline{1-1} \cline{3-3} \cline{5-5} \cline{7-7} 
		All features-ELM &  & 0.4653 &  & 0.3582 &  & 0.6624 \\
		All features-SVR &  & 0.6519 &  & 0.4986 &  & 0.3949 \\
		GGA-ELM-ELM      &  & 0.3650 &  & 0.2858 &  & 0.7049 \\
		GGA-ELM-SVR      &  & 0.3599 &  & 0.2727 &  & 0.7056 \\
		BO-GGA-ELM-ELM   &  & 0.3324 &  & 0.2519 &  & 0.7429 \\
		BO-GGA-ELM-SVR   &  & 0.3331 &  & 0.2461 &  & 0.7396 \\
		Seqtoseq (Adam)  &  & 0.3419 &  & 0.2528 &  & {\bf0.8106}  \\
		LSTM+FCL (Adam)  &  & {\bf0.2784} &  & {\bf0.2392} &  & 0.7922 \\ \hline
	\end{tabular}
\end{table}

From the table, it is clear that the performance of the seqtoseq network is very promising, as it obtains the best $r^2$ values among the methods. Moreover, LSTM+FCL significantly outperforms all the alternatives and even the seqtoseq network in terms of RMSE and MAE simultaneously. It is worth mentioning that both of our methods have access only to $H_s$ of the stations, while the benchmark methods use more measurements. For example, the GAA methods obtain their best performance using wind speed, significant wave height, average wave period, air temperature and the atmospheric pressure at buoy 46025 and the significant wave height and average wave period of buoy 46042 \cite{cornejo2016significant}. This makes the superiority of our methods even clearer, but brings about another question, as well, namely: Would using proper feature selection improve the performance of deep networks even further? We conducted some preliminary experiments to answer this question, considering combinations of the features that the GAA approaches use,  and found an inferior accuracy compared to solely using $H_s$. In Section~\ref{sec:feature_selection}, we discuss a more efficient approach for feature selection. 

\begin{figure}[htbp]
	\hspace*{-2cm}
	\scalebox{0.42}{\input{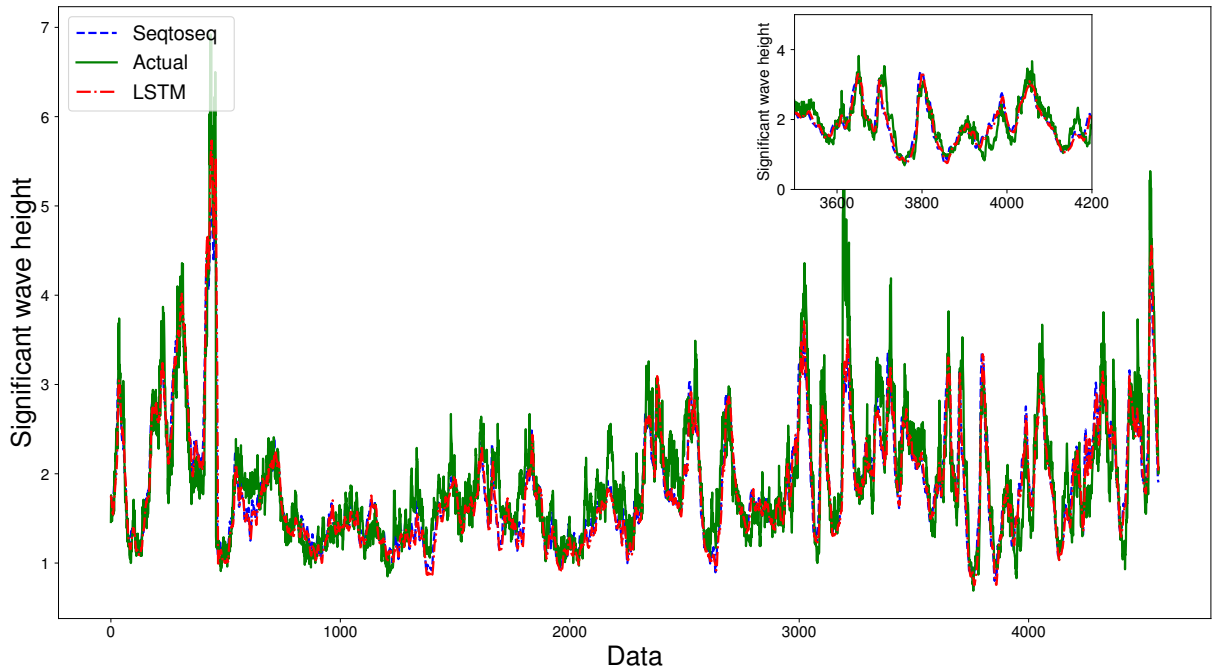}}
	\caption{Significant wave height reconstruction of buoy 46069. Inset shows a zoomed-in portion.}
	\label{fig:2}
\end{figure}

Figure \ref{fig:2} illustrates the performance of both the LSTM and seqtoseq networks in reconstructing the significant wave height for buoy 46069. Note that the models never ``saw'' the solid green line, which is the year 2010 data (i.e., the test data).

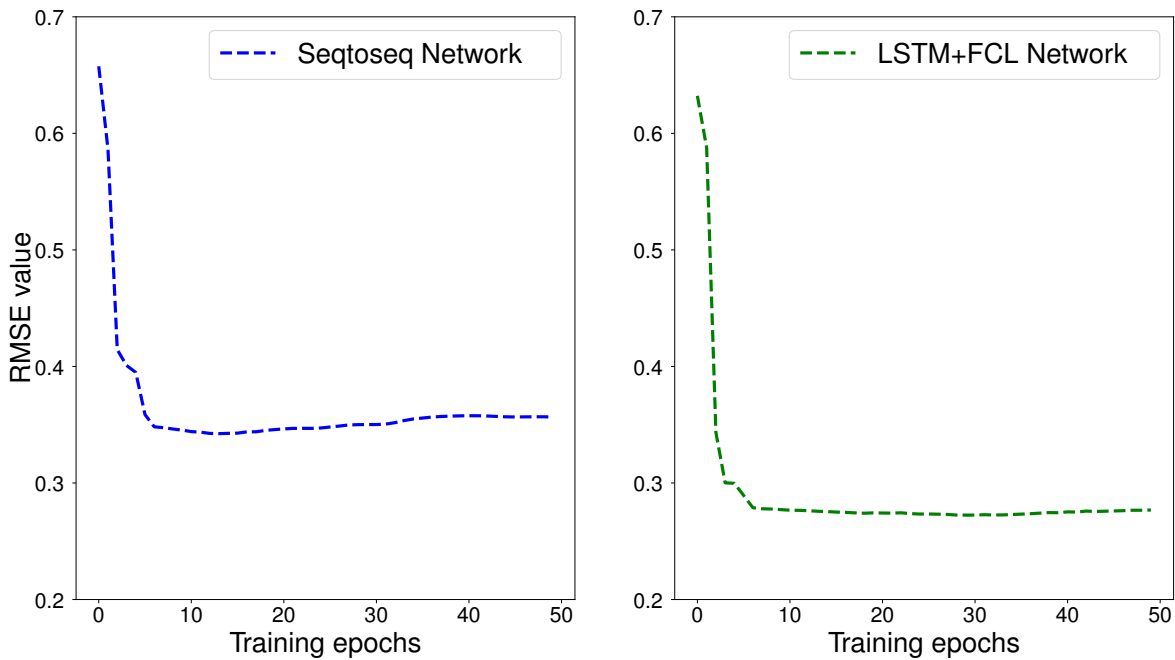
\begin{figure}[htbp]
	\hspace*{-1.25cm}
		\scalebox{0.4}{\input{testRMSE1.pgf}}
	\caption{$H_s$ RMSE test (year 2010) behavior on training data (year 2009) of buoy 46069.}
	\label{fig:3}
\end{figure}

We also illustrate the RMSE behavior of LSTM+FCL and seqtoseq. Figure \ref{fig:3} displays the RMSE test errors based on the number of training epochs up to 50. 
Both networks reach their best performance around tenth epoch with sequence-to-sequence reaching slightly faster. This happens even though there are more weights in the sequence-to-sequence model. We argue that epoch-scheduling training may be responsible for this quicker training performance, because by definition it allows the network to see ground truth outputs to update the parameters better while simultaneously avoiding excessive overfittings.

\section{Feature Selection}
\label{sec:feature_selection}

In this section, we modify the proposed framework to investigate feature selection via elastic net \cite{zou2005regularization} methodology. We alter the loss function to include two separate regularizers, ``$\ell_1$ norm'' and ``$\ell_2$ norm'', as follows:   
$$ f_w(x)= \frac{1}{n} \sum_{i=1}^{n} \left(  y^i -N(x^i;w)  \right) ^2 + \lambda_1 \|w \|^2_2 + \lambda_2 \|w \|_1$$
where $ \|w \|_1= \sum_{W} w_i$ and $W$ is the set of all the network weights. The ``$\ell_1$ norm'' inherits a naturally sparse collection of variables. $\|w \|_1$ has its minimum when all the weights are zero (uncollected) so every single nonzero weight should improve the first term of the objective function. Hence, the model chooses only those weights for which the decrease in the first term outweighs the increase in the last term. Using only the ``$\ell_1$ norm'' regularizer would result in instability among the multiple solutions because a tiny change in the parameters may change some weights from zero to nonzero values or vice versa \cite{ng2004feature}. Therefore, the ``$\ell_2$ norm'' is used in the objective.

One important difference between our feature selection method and those used in the wave energy literature (e.g., \cite{cornejo2018bayesian}) is that the models in the literature either use all of the data related to a specific feature, like $H_s$, from a nearby buoy, or decide to ignore that feature entirely; however, our model allows us to partially utilize any of the features based on nonzero weights resulting in the optimal solution.

\begin{figure}[htbp]
	\centering
	\includegraphics[width = 9cm]{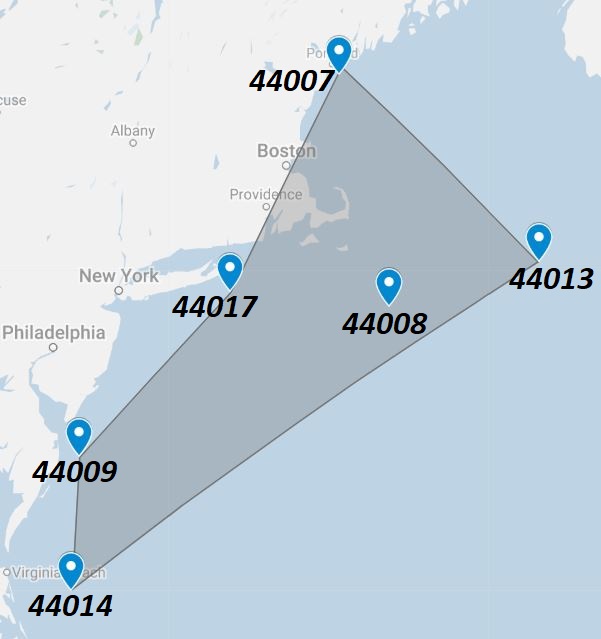}
	\caption{Adjacent buoys of 44007, 44008, 44009, 44013, 44014 and 44017}
	\label{fig:buoysFS}
\end{figure}

To investigate the performance of our proposed framework for feature collection, we design a new experiment which has been recently explored for feature selection via ensemble structures as well \cite{pirhooshyaran2020feature}. Figure \ref{fig:buoysFS} illustrates six stations---44007, 44008, 44009, 44013, 44014 and 44017---from the east coast of the United States. To see their exact locations and water depths, you can refer to Table \ref{tab1}. We reconstruct the significant wave height  and power output of buoy 44008 based on available features of the other nearby buoys. We chose these buoys for two main reasons. First,  on average they are rich in features with usable data. Second, the resolution and the data-collection times for these stations are the same up to 1-minute accuracy. This is crucial when we want to reconstruct a feature of a buoy based on others. We consider wind direction (WDIR), wind speed (WSPD), gust speed (GST), significant wave height (WVHT), dominant wave period (DPD), average wave period (APD), direction of dominant period waves (MWD), sea level pressure (PRES), air temperature (ATMP), sea surface temperature (WTMP) and dewpoint temperature (DEWP). The datasets are for the year 2018. We include all the features for all the buoys, except DEWP for buoys 44009 and 44013 because those data are unusable. We exclude a measurement for all the stations only if a feature is missing in at least one of them. In the end, the dataset consists of 3422 measurements of 53 features. We divide the dataset into train (60\%), validation (20\%) and test (20\%). We do not change the real measurements in any way. In addition, the methods never ``see'' the test set during training and validation. In this experiment, for both the LSTM+FCL and seqtoseq networks we have one more parameter to tune, namely, the coefficient of $\mathbb{L}$1-norm) during the validation.      

Table \ref{tab7} compares the performance of all of the methods, along with an alternative method, Random Forest, for reconstructing both $H_s$ and $\mathcal{P}$ wave features. There is a column dedicated to the percentage of non-zero trainable variables. We consider a single weight or bias variable to be non-zero if its absolute value exceeds the threshold of $0.0001$, and we report the result as the percentage of non-zero variables among all trainable variables. When we use the elastic net concept for selecting weights and biases, we observe a significant drop in the non-zero values. This  means exactly what we discussed at the beginning of this section. The models invest only in those variables which are beneficial to the objective value.  
Here, we report the best results found for each measurement error RMSE, HUBER and MAAPE independently. In other words, the reported values of RMSE, HUBER or MAAPE are not necessarily gathered from a single experiment, but instead they are the best ones for the training epochs less than or equal to 50. As seen in Table \ref{tab7}, EN generally improves the performance of each method. For instance, EN aids single-layered LSTM+FCL significantly for reconstructing $H_s$ and reduces RMSE and HUBER losses by 29\% and 54\%, respectively. Similarly, it helps the seqtoseq network slightly and reduces the measurement errors by around 5\%. \\
Another interesting finding is the comparison between multi-layered LSTM+FCL and single-layered ones. Unlike in the prediction section \ref{sec:network_comparison}, here the performance of multi-layered LSTM+FCL is satisfactory, and even superior in many cases. Therefore, we argue that for feature selection, when the number of features is considerable compared to forecasting, having deeper structures improves the performance.

In addition, we observe that networks with more variables, such as multi-layered LSTM+FCL or seqtoseq, retain more non-zeros. In particular, in the process of tuning, they choose the $\mathbb{L}$1-norm coefficient to be smaller. Hence, the effect of the $\mathbb{L}$1-norm is lessened. This may be due to the fact that the more advanced structures can find hidden relations among features and set their corresponding weights to non-zero values, whereas single-layered models are unable to discover some hidden relations, so they set the weights equal to zero.\\
The Random Forest algorithm \cite{breiman2001random} is known to be very efficient and accurate in ranking the importance of variables. To have a fair comparison, we hypertune its major parameters such as maximum depth. For $H_s$ reconstruction, it finds the best RMSE and HUBER losses, but for $\mathcal{P}$ reconstruction, it fails to provide similar performance. The seqtoseq+EN framework is the most robust and promising one. It attains the best performance for half the measurements and is generally among the top methods for the rest. We observe that $\lambda_i \in [0.001,0.01], \ i=1,2$ results in the leading performance for the seqtoseq network.

\begin{table}[]
			\caption{ Comparison of different methods reconstructing $H_s$ and $\mathcal{P}$ of 44008 station, epochs $\leq 50$.}
	\label{tab7}
		\setlength\tabcolsep{3pt}
		\centering
	\begin{tabular}{lllllllllll}
		&  & \multicolumn{9}{c}{\textbf{Wave Feature}}                                                                                                                                                                                                                                     \\ \cline{3-11} 
		&  & \multicolumn{4}{c}{\textbf{$H_s$}}                                                                                                      &  & \multicolumn{4}{c}{\textbf{$\mathcal{P}$}}                                                                                                      \\ \cline{3-6} \cline{8-11} 
		Methods        &  & RMSE (m)       & HUBER          & MAAPE          & \multicolumn{1}{c}{\begin{tabular}[c]{@{}c@{}}Non-Zero  \\  Var(\%)\end{tabular}} &  & RMSE (m)       & HUBER          & MAAPE          & \multicolumn{1}{c}{\begin{tabular}[c]{@{}c@{}}Non-Zero \\  Var(\%)\end{tabular}} \\ \cline{1-1} \cline{3-6} \cline{8-11} 
		SL-LSTM+FCL         &  & 0.569          & 0.209          & 0.301          & 99.4                                                                              &  & 3.031          & 8.394          & 0.520          & 99.2                                                                             \\
		ML-LSTM+FCL         &  & 0.391          & 0.120          & 0.304          & 99.7                                                                              &  & 3.065          & 9.041          & 0.510          & 99.9                                                                             \\
		Seqtoseq       &  & 0.469          & 0.108          & 0.279          & 99.7                                                                              &  & 4.593          & 2.229          & 0.486          & 99.7                                                                             \\
		SL-LSTM+FCL +EN     &  & 0.439          & 0.135          & 0.354          & 20.0                                                                              &  & 3.171          & 7.543          & 0.500          & 16.2                                                                             \\
		ML-LSTM+FCL +EN     &  & 0.368          & 0.110          & 0.289          & 33.6                                                                              &  & \textbf{2.993} & 9.224          & 0.500          & 50.7                                                                             \\
		Seqtoseq + EN  &  & 0.459          & 0.106          & \textbf{0.264} & 58.2                                                                              &  & 4.526          & \textbf{2.211} & \textbf{0.455} & 69.8                                                                             \\
		Random Forrest &  & \textbf{0.359} & \textbf{0.055} & 0.269          & \multicolumn{1}{c}{}                                                              &  & 3.431          & 5.618          & 0.561          & \multicolumn{1}{c}{}                                                             \\ \cline{1-11}  
	\end{tabular}
\end{table}

\section{Conclusion} 
\label{conclu}
This article focuses on the reconstruction, feature selection and multivariate, multistep forecasting of ocean wave characteristics based on real data obtained from NOAA buoy measurements around the globe. This paper is among the first to propose the use of sequence-to-sequence and other novel recurrent networks for these objectives. Moreover, the epoch-scheduled training concept has been introduced as a soft technique so that the model enjoys the consistency of teacher forcing methods while avoiding overfitting. We tested various optimization algorithms on the networks that we introduced.  AMSGrad and Adam present robust and promising performance comparing with SGD and RMSProp. We compared several recurrent networks. All of the parameters are tuned with Spearmint Bayesian optimization under the same budget. One observes that Spearmint favors single-layered networks as opposed to multi-layered networks for multivariate forecasting. This can be observed in our numerical studies as well. The SL-LSTM+FCL and seqtoseq models demonstrate superior performance compared to the other networks.

Furthermore, we explore the problem of reconstructing wave features, which has been well studied in the literature. The results suggest the superiority of our proposed networks compared to existing techniques. Our findings emphasize the fact that using only significant wave heights of adjacent buoys is sufficient to construct $H_s$ of a nearby station. 
Proper feature selection, however, necessitates more analysis. We design a new experiment using NOAA data from the east coast of the United States. The dataset can be used for any feature selection and multivariate regression research. We incorporated the elastic net concept into our proposed neural networks to handle 53 features of the dataset. We evaluate the performance with and without the EN and with the random forest algorithm. The results indicate that the seqtoseq network has a consistent and reliable performance. In addition, for feature selection, deeper recurrent structures are more promising compared to single-layered ones. 

We suggest the use of different parameter tuning algorithms such as derivative-free optimization instead of Spearmint as one promising future direction. Moreover, utilizing other neural network techniques---such as attention mechanism, in particular---for wave feature selection and reconstruction can be another area for future study.    

{\bf Acknowledgments}\\
This research was partially supported by the National Science Foundation (NSF) through the CyberSEES grant \#1442858. The authors are responsible for the results presented in the paper. These results have not been endorsed by the sponsoring agency.  			            	

\bibliography{forecastinPaper}

\bibliographystyle{unsrtnat}

\end{document}

%% file: hyperParameter.tex
	
\tikzstyle{decision} = [diamond,   draw, fill=white!20, text width=6em,  text centered,  node distance=3cm, inner sep=0pt]
\tikzstyle{block0}   = [rectangle, draw, fill=white!20, text width=25em, text centered,  node distance=3cm, rounded corners, minimum height=4em]
\tikzstyle{block00}  = [rectangle, draw, fill=white!20, text width=35em, text centered,  node distance=3cm, rounded corners, minimum height=4em]
\tikzstyle{block}    = [rectangle, draw, fill=white!20, text width=9em,  text centered,                     rounded corners, minimum height=4em]
\tikzstyle{block1}   = [rectangle, draw, fill=white!20, text width=16em, text centered,                     rounded corners, minimum height=4em]
\tikzstyle{line}     = [draw, -latex']

  \centering

  \begin{tikzpicture}[node distance = 2cm, auto]
    \node [block0, align=center] (init) { \begin{itemize} \item Define the black-box function $f(x)$ \item Define input parameters and their feasible intervals \item Define number of function evaluations (budget)\end{itemize}};
    \node [block00, align=center,below of = init] (GP) { {\bf Gaussian Process (GP)}: \\ For $(x,y)$ observation, assume $y\sim \mathcal{N}(f(x),\nu)$ where $\nu$ is the noise variance};
    \node [block, below of = GP,node distance = 2.5cm] (input parameters) {Initialize input parameters};
    \node [decision, below of = input parameters] (Budget) { Budget {\bf Left}?};
    \node [block, right of = Budget, node distance = 4cm] (no block) {Store the best parameters};
    \node [block, left of = Budget, node distance = 4cm] (yes block1) {Train the network \\({\bf loss}: MSE+L2)};
    \node [block, below of = yes block1, node distance = 2cm] (yes block2) {Find the prediction};
    \node [block, below of = yes block2, node distance = 2cm] (yes block3) {Find the black-box function output \\ ({\bf loss}: MAAPE)};
    \node [block1, below of = yes block3, node distance = 2cm] (yes block4) {{\bf Expected Improvement (EI)}:\\ Find the next best feasible point};
    \node [block, below of = Budget, node distance = 8cm] (best pred) {Find the best prediction};
    \node [block, left of = yes block1, node distance = 5cm] (Tr) {Train};
	\node [block, left of = yes block2, node distance = 5cm] (Val) {Validation};
	\node [block, below of = Val, node distance = 6cm] (Te) {Test};
	\node [block, left of = yes block3, node distance = 8cm] (Da) {Data};
	\node [block1, below of = best pred, node distance = 2cm] (Re) {Report the error metrics};
    \path [line] (init) -- (GP);
    \path [line] (GP) -- (input parameters);
    \path [line] (input parameters) -- (Budget);
    \path [line] (Budget) -- node { Yes}(yes block1);
    \path [line] (Budget) -- node { No}(no block);
    \path [line] (yes block1) -- node {}(yes block2);
    \path [line] (yes block2) -- node {}(yes block3);    
    \path [line] (yes block3) -- node {}(yes block4);    
    \path [line] (yes block4) -| node {}(Budget);        
    \path [line] (Tr) -- node {}(yes block1);
    \path [line] (Val) -- node {}(yes block2);
    \path [line] (Te) -- node {}(best pred); 
    \path [line] (no block) |- node {}(best pred);   
    \path [line] (Da) |- node {}(Tr);      
    \path [line] (Da) |- node {}(Val);       
    \path [line] (Da) |- node {}(Te); 
    \path [line] (best pred) -- node {}(Re);                          
  \end{tikzpicture}

%% file: testRMSE1.pgf
\begingroup%
\makeatletter%
\begin{pgfpicture}%
\pgfpathrectangle{\pgfpointorigin}{\pgfqpoint{18.530000in}{9.910000in}}%
\pgfusepath{use as bounding box, clip}%
\begin{pgfscope}%
\pgfsetbuttcap%
\pgfsetmiterjoin%
\definecolor{currentfill}{rgb}{1.000000,1.000000,1.000000}%
\pgfsetfillcolor{currentfill}%
\pgfsetlinewidth{0.000000pt}%
\definecolor{currentstroke}{rgb}{1.000000,1.000000,1.000000}%
\pgfsetstrokecolor{currentstroke}%
\pgfsetdash{}{0pt}%
\pgfpathmoveto{\pgfqpoint{0.000000in}{0.000000in}}%
\pgfpathlineto{\pgfqpoint{18.530000in}{0.000000in}}%
\pgfpathlineto{\pgfqpoint{18.530000in}{9.910000in}}%
\pgfpathlineto{\pgfqpoint{0.000000in}{9.910000in}}%
\pgfpathclose%
\pgfusepath{fill}%
\end{pgfscope}%
\begin{pgfscope}%
\pgfsetbuttcap%
\pgfsetmiterjoin%
\definecolor{currentfill}{rgb}{1.000000,1.000000,1.000000}%
\pgfsetfillcolor{currentfill}%
\pgfsetlinewidth{0.000000pt}%
\definecolor{currentstroke}{rgb}{0.000000,0.000000,0.000000}%
\pgfsetstrokecolor{currentstroke}%
\pgfsetstrokeopacity{0.000000}%
\pgfsetdash{}{0pt}%
\pgfpathmoveto{\pgfqpoint{2.316250in}{1.090100in}}%
\pgfpathlineto{\pgfqpoint{8.843864in}{1.090100in}}%
\pgfpathlineto{\pgfqpoint{8.843864in}{8.720800in}}%
\pgfpathlineto{\pgfqpoint{2.316250in}{8.720800in}}%
\pgfpathclose%
\pgfusepath{fill}%
\end{pgfscope}%
\begin{pgfscope}%
\pgfsetbuttcap%
\pgfsetroundjoin%
\definecolor{currentfill}{rgb}{0.000000,0.000000,0.000000}%
\pgfsetfillcolor{currentfill}%
\pgfsetlinewidth{0.803000pt}%
\definecolor{currentstroke}{rgb}{0.000000,0.000000,0.000000}%
\pgfsetstrokecolor{currentstroke}%
\pgfsetdash{}{0pt}%
\pgfsys@defobject{currentmarker}{\pgfqpoint{0.000000in}{-0.048611in}}{\pgfqpoint{0.000000in}{0.000000in}}{%
\pgfpathmoveto{\pgfqpoint{0.000000in}{0.000000in}}%
\pgfpathlineto{\pgfqpoint{0.000000in}{-0.048611in}}%
\pgfusepath{stroke,fill}%
}%
\begin{pgfscope}%
\pgfsys@transformshift{2.612960in}{1.090100in}%
\pgfsys@useobject{currentmarker}{}%
\end{pgfscope}%
\end{pgfscope}%
\begin{pgfscope}%
\definecolor{textcolor}{rgb}{0.000000,0.000000,0.000000}%
\pgfsetstrokecolor{textcolor}%
\pgfsetfillcolor{textcolor}%
\pgftext[x=2.612960in,y=0.992878in,,top]{\color{textcolor}\sffamily\fontsize{20.000000}{24.000000}\selectfont 0}%
\end{pgfscope}%
\begin{pgfscope}%
\pgfsetbuttcap%
\pgfsetroundjoin%
\definecolor{currentfill}{rgb}{0.000000,0.000000,0.000000}%
\pgfsetfillcolor{currentfill}%
\pgfsetlinewidth{0.803000pt}%
\definecolor{currentstroke}{rgb}{0.000000,0.000000,0.000000}%
\pgfsetstrokecolor{currentstroke}%
\pgfsetdash{}{0pt}%
\pgfsys@defobject{currentmarker}{\pgfqpoint{0.000000in}{-0.048611in}}{\pgfqpoint{0.000000in}{0.000000in}}{%
\pgfpathmoveto{\pgfqpoint{0.000000in}{0.000000in}}%
\pgfpathlineto{\pgfqpoint{0.000000in}{-0.048611in}}%
\pgfusepath{stroke,fill}%
}%
\begin{pgfscope}%
\pgfsys@transformshift{3.824020in}{1.090100in}%
\pgfsys@useobject{currentmarker}{}%
\end{pgfscope}%
\end{pgfscope}%
\begin{pgfscope}%
\definecolor{textcolor}{rgb}{0.000000,0.000000,0.000000}%
\pgfsetstrokecolor{textcolor}%
\pgfsetfillcolor{textcolor}%
\pgftext[x=3.824020in,y=0.992878in,,top]{\color{textcolor}\sffamily\fontsize{20.000000}{24.000000}\selectfont 10}%
\end{pgfscope}%
\begin{pgfscope}%
\pgfsetbuttcap%
\pgfsetroundjoin%
\definecolor{currentfill}{rgb}{0.000000,0.000000,0.000000}%
\pgfsetfillcolor{currentfill}%
\pgfsetlinewidth{0.803000pt}%
\definecolor{currentstroke}{rgb}{0.000000,0.000000,0.000000}%
\pgfsetstrokecolor{currentstroke}%
\pgfsetdash{}{0pt}%
\pgfsys@defobject{currentmarker}{\pgfqpoint{0.000000in}{-0.048611in}}{\pgfqpoint{0.000000in}{0.000000in}}{%
\pgfpathmoveto{\pgfqpoint{0.000000in}{0.000000in}}%
\pgfpathlineto{\pgfqpoint{0.000000in}{-0.048611in}}%
\pgfusepath{stroke,fill}%
}%
\begin{pgfscope}%
\pgfsys@transformshift{5.035080in}{1.090100in}%
\pgfsys@useobject{currentmarker}{}%
\end{pgfscope}%
\end{pgfscope}%
\begin{pgfscope}%
\definecolor{textcolor}{rgb}{0.000000,0.000000,0.000000}%
\pgfsetstrokecolor{textcolor}%
\pgfsetfillcolor{textcolor}%
\pgftext[x=5.035080in,y=0.992878in,,top]{\color{textcolor}\sffamily\fontsize{20.000000}{24.000000}\selectfont 20}%
\end{pgfscope}%
\begin{pgfscope}%
\pgfsetbuttcap%
\pgfsetroundjoin%
\definecolor{currentfill}{rgb}{0.000000,0.000000,0.000000}%
\pgfsetfillcolor{currentfill}%
\pgfsetlinewidth{0.803000pt}%
\definecolor{currentstroke}{rgb}{0.000000,0.000000,0.000000}%
\pgfsetstrokecolor{currentstroke}%
\pgfsetdash{}{0pt}%
\pgfsys@defobject{currentmarker}{\pgfqpoint{0.000000in}{-0.048611in}}{\pgfqpoint{0.000000in}{0.000000in}}{%
\pgfpathmoveto{\pgfqpoint{0.000000in}{0.000000in}}%
\pgfpathlineto{\pgfqpoint{0.000000in}{-0.048611in}}%
\pgfusepath{stroke,fill}%
}%
\begin{pgfscope}%
\pgfsys@transformshift{6.246140in}{1.090100in}%
\pgfsys@useobject{currentmarker}{}%
\end{pgfscope}%
\end{pgfscope}%
\begin{pgfscope}%
\definecolor{textcolor}{rgb}{0.000000,0.000000,0.000000}%
\pgfsetstrokecolor{textcolor}%
\pgfsetfillcolor{textcolor}%
\pgftext[x=6.246140in,y=0.992878in,,top]{\color{textcolor}\sffamily\fontsize{20.000000}{24.000000}\selectfont 30}%
\end{pgfscope}%
\begin{pgfscope}%
\pgfsetbuttcap%
\pgfsetroundjoin%
\definecolor{currentfill}{rgb}{0.000000,0.000000,0.000000}%
\pgfsetfillcolor{currentfill}%
\pgfsetlinewidth{0.803000pt}%
\definecolor{currentstroke}{rgb}{0.000000,0.000000,0.000000}%
\pgfsetstrokecolor{currentstroke}%
\pgfsetdash{}{0pt}%
\pgfsys@defobject{currentmarker}{\pgfqpoint{0.000000in}{-0.048611in}}{\pgfqpoint{0.000000in}{0.000000in}}{%
\pgfpathmoveto{\pgfqpoint{0.000000in}{0.000000in}}%
\pgfpathlineto{\pgfqpoint{0.000000in}{-0.048611in}}%
\pgfusepath{stroke,fill}%
}%
\begin{pgfscope}%
\pgfsys@transformshift{7.457200in}{1.090100in}%
\pgfsys@useobject{currentmarker}{}%
\end{pgfscope}%
\end{pgfscope}%
\begin{pgfscope}%
\definecolor{textcolor}{rgb}{0.000000,0.000000,0.000000}%
\pgfsetstrokecolor{textcolor}%
\pgfsetfillcolor{textcolor}%
\pgftext[x=7.457200in,y=0.992878in,,top]{\color{textcolor}\sffamily\fontsize{20.000000}{24.000000}\selectfont 40}%
\end{pgfscope}%
\begin{pgfscope}%
\pgfsetbuttcap%
\pgfsetroundjoin%
\definecolor{currentfill}{rgb}{0.000000,0.000000,0.000000}%
\pgfsetfillcolor{currentfill}%
\pgfsetlinewidth{0.803000pt}%
\definecolor{currentstroke}{rgb}{0.000000,0.000000,0.000000}%
\pgfsetstrokecolor{currentstroke}%
\pgfsetdash{}{0pt}%
\pgfsys@defobject{currentmarker}{\pgfqpoint{0.000000in}{-0.048611in}}{\pgfqpoint{0.000000in}{0.000000in}}{%
\pgfpathmoveto{\pgfqpoint{0.000000in}{0.000000in}}%
\pgfpathlineto{\pgfqpoint{0.000000in}{-0.048611in}}%
\pgfusepath{stroke,fill}%
}%
\begin{pgfscope}%
\pgfsys@transformshift{8.668260in}{1.090100in}%
\pgfsys@useobject{currentmarker}{}%
\end{pgfscope}%
\end{pgfscope}%
\begin{pgfscope}%
\definecolor{textcolor}{rgb}{0.000000,0.000000,0.000000}%
\pgfsetstrokecolor{textcolor}%
\pgfsetfillcolor{textcolor}%
\pgftext[x=8.668260in,y=0.992878in,,top]{\color{textcolor}\sffamily\fontsize{20.000000}{24.000000}\selectfont 50}%
\end{pgfscope}%
\begin{pgfscope}%
\definecolor{textcolor}{rgb}{0.000000,0.000000,0.000000}%
\pgfsetstrokecolor{textcolor}%
\pgfsetfillcolor{textcolor}%
\pgftext[x=5.580057in,y=0.668497in,,top]{\color{textcolor}\sffamily\fontsize{26.000000}{31.200000}\selectfont Training epochs}%
\end{pgfscope}%
\begin{pgfscope}%
\pgfsetbuttcap%
\pgfsetroundjoin%
\definecolor{currentfill}{rgb}{0.000000,0.000000,0.000000}%
\pgfsetfillcolor{currentfill}%
\pgfsetlinewidth{0.803000pt}%
\definecolor{currentstroke}{rgb}{0.000000,0.000000,0.000000}%
\pgfsetstrokecolor{currentstroke}%
\pgfsetdash{}{0pt}%
\pgfsys@defobject{currentmarker}{\pgfqpoint{-0.048611in}{0.000000in}}{\pgfqpoint{0.000000in}{0.000000in}}{%
\pgfpathmoveto{\pgfqpoint{0.000000in}{0.000000in}}%
\pgfpathlineto{\pgfqpoint{-0.048611in}{0.000000in}}%
\pgfusepath{stroke,fill}%
}%
\begin{pgfscope}%
\pgfsys@transformshift{2.316250in}{1.090100in}%
\pgfsys@useobject{currentmarker}{}%
\end{pgfscope}%
\end{pgfscope}%
\begin{pgfscope}%
\definecolor{textcolor}{rgb}{0.000000,0.000000,0.000000}%
\pgfsetstrokecolor{textcolor}%
\pgfsetfillcolor{textcolor}%
\pgftext[x=1.777269in,y=0.984577in,left,base]{\color{textcolor}\sffamily\fontsize{20.000000}{24.000000}\selectfont 0.2}%
\end{pgfscope}%
\begin{pgfscope}%
\pgfsetbuttcap%
\pgfsetroundjoin%
\definecolor{currentfill}{rgb}{0.000000,0.000000,0.000000}%
\pgfsetfillcolor{currentfill}%
\pgfsetlinewidth{0.803000pt}%
\definecolor{currentstroke}{rgb}{0.000000,0.000000,0.000000}%
\pgfsetstrokecolor{currentstroke}%
\pgfsetdash{}{0pt}%
\pgfsys@defobject{currentmarker}{\pgfqpoint{-0.048611in}{0.000000in}}{\pgfqpoint{0.000000in}{0.000000in}}{%
\pgfpathmoveto{\pgfqpoint{0.000000in}{0.000000in}}%
\pgfpathlineto{\pgfqpoint{-0.048611in}{0.000000in}}%
\pgfusepath{stroke,fill}%
}%
\begin{pgfscope}%
\pgfsys@transformshift{2.316250in}{2.616240in}%
\pgfsys@useobject{currentmarker}{}%
\end{pgfscope}%
\end{pgfscope}%
\begin{pgfscope}%
\definecolor{textcolor}{rgb}{0.000000,0.000000,0.000000}%
\pgfsetstrokecolor{textcolor}%
\pgfsetfillcolor{textcolor}%
\pgftext[x=1.777269in,y=2.510717in,left,base]{\color{textcolor}\sffamily\fontsize{20.000000}{24.000000}\selectfont 0.3}%
\end{pgfscope}%
\begin{pgfscope}%
\pgfsetbuttcap%
\pgfsetroundjoin%
\definecolor{currentfill}{rgb}{0.000000,0.000000,0.000000}%
\pgfsetfillcolor{currentfill}%
\pgfsetlinewidth{0.803000pt}%
\definecolor{currentstroke}{rgb}{0.000000,0.000000,0.000000}%
\pgfsetstrokecolor{currentstroke}%
\pgfsetdash{}{0pt}%
\pgfsys@defobject{currentmarker}{\pgfqpoint{-0.048611in}{0.000000in}}{\pgfqpoint{0.000000in}{0.000000in}}{%
\pgfpathmoveto{\pgfqpoint{0.000000in}{0.000000in}}%
\pgfpathlineto{\pgfqpoint{-0.048611in}{0.000000in}}%
\pgfusepath{stroke,fill}%
}%
\begin{pgfscope}%
\pgfsys@transformshift{2.316250in}{4.142380in}%
\pgfsys@useobject{currentmarker}{}%
\end{pgfscope}%
\end{pgfscope}%
\begin{pgfscope}%
\definecolor{textcolor}{rgb}{0.000000,0.000000,0.000000}%
\pgfsetstrokecolor{textcolor}%
\pgfsetfillcolor{textcolor}%
\pgftext[x=1.777269in,y=4.036857in,left,base]{\color{textcolor}\sffamily\fontsize{20.000000}{24.000000}\selectfont 0.4}%
\end{pgfscope}%
\begin{pgfscope}%
\pgfsetbuttcap%
\pgfsetroundjoin%
\definecolor{currentfill}{rgb}{0.000000,0.000000,0.000000}%
\pgfsetfillcolor{currentfill}%
\pgfsetlinewidth{0.803000pt}%
\definecolor{currentstroke}{rgb}{0.000000,0.000000,0.000000}%
\pgfsetstrokecolor{currentstroke}%
\pgfsetdash{}{0pt}%
\pgfsys@defobject{currentmarker}{\pgfqpoint{-0.048611in}{0.000000in}}{\pgfqpoint{0.000000in}{0.000000in}}{%
\pgfpathmoveto{\pgfqpoint{0.000000in}{0.000000in}}%
\pgfpathlineto{\pgfqpoint{-0.048611in}{0.000000in}}%
\pgfusepath{stroke,fill}%
}%
\begin{pgfscope}%
\pgfsys@transformshift{2.316250in}{5.668520in}%
\pgfsys@useobject{currentmarker}{}%
\end{pgfscope}%
\end{pgfscope}%
\begin{pgfscope}%
\definecolor{textcolor}{rgb}{0.000000,0.000000,0.000000}%
\pgfsetstrokecolor{textcolor}%
\pgfsetfillcolor{textcolor}%
\pgftext[x=1.777269in,y=5.562997in,left,base]{\color{textcolor}\sffamily\fontsize{20.000000}{24.000000}\selectfont 0.5}%
\end{pgfscope}%
\begin{pgfscope}%
\pgfsetbuttcap%
\pgfsetroundjoin%
\definecolor{currentfill}{rgb}{0.000000,0.000000,0.000000}%
\pgfsetfillcolor{currentfill}%
\pgfsetlinewidth{0.803000pt}%
\definecolor{currentstroke}{rgb}{0.000000,0.000000,0.000000}%
\pgfsetstrokecolor{currentstroke}%
\pgfsetdash{}{0pt}%
\pgfsys@defobject{currentmarker}{\pgfqpoint{-0.048611in}{0.000000in}}{\pgfqpoint{0.000000in}{0.000000in}}{%
\pgfpathmoveto{\pgfqpoint{0.000000in}{0.000000in}}%
\pgfpathlineto{\pgfqpoint{-0.048611in}{0.000000in}}%
\pgfusepath{stroke,fill}%
}%
\begin{pgfscope}%
\pgfsys@transformshift{2.316250in}{7.194660in}%
\pgfsys@useobject{currentmarker}{}%
\end{pgfscope}%
\end{pgfscope}%
\begin{pgfscope}%
\definecolor{textcolor}{rgb}{0.000000,0.000000,0.000000}%
\pgfsetstrokecolor{textcolor}%
\pgfsetfillcolor{textcolor}%
\pgftext[x=1.777269in,y=7.089137in,left,base]{\color{textcolor}\sffamily\fontsize{20.000000}{24.000000}\selectfont 0.6}%
\end{pgfscope}%
\begin{pgfscope}%
\pgfsetbuttcap%
\pgfsetroundjoin%
\definecolor{currentfill}{rgb}{0.000000,0.000000,0.000000}%
\pgfsetfillcolor{currentfill}%
\pgfsetlinewidth{0.803000pt}%
\definecolor{currentstroke}{rgb}{0.000000,0.000000,0.000000}%
\pgfsetstrokecolor{currentstroke}%
\pgfsetdash{}{0pt}%
\pgfsys@defobject{currentmarker}{\pgfqpoint{-0.048611in}{0.000000in}}{\pgfqpoint{0.000000in}{0.000000in}}{%
\pgfpathmoveto{\pgfqpoint{0.000000in}{0.000000in}}%
\pgfpathlineto{\pgfqpoint{-0.048611in}{0.000000in}}%
\pgfusepath{stroke,fill}%
}%
\begin{pgfscope}%
\pgfsys@transformshift{2.316250in}{8.720800in}%
\pgfsys@useobject{currentmarker}{}%
\end{pgfscope}%
\end{pgfscope}%
\begin{pgfscope}%
\definecolor{textcolor}{rgb}{0.000000,0.000000,0.000000}%
\pgfsetstrokecolor{textcolor}%
\pgfsetfillcolor{textcolor}%
\pgftext[x=1.777269in,y=8.615277in,left,base]{\color{textcolor}\sffamily\fontsize{20.000000}{24.000000}\selectfont 0.7}%
\end{pgfscope}%
\begin{pgfscope}%
\definecolor{textcolor}{rgb}{0.000000,0.000000,0.000000}%
\pgfsetstrokecolor{textcolor}%
\pgfsetfillcolor{textcolor}%
\pgftext[x=1.721713in,y=4.905450in,,bottom,rotate=90.000000]{\color{textcolor}\sffamily\fontsize{26.000000}{31.200000}\selectfont RMSE value}%
\end{pgfscope}%
\begin{pgfscope}%
\pgfpathrectangle{\pgfqpoint{2.316250in}{1.090100in}}{\pgfqpoint{6.527614in}{7.630700in}}%
\pgfusepath{clip}%
\pgfsetbuttcap%
\pgfsetroundjoin%
\pgfsetlinewidth{3.011250pt}%
\definecolor{currentstroke}{rgb}{0.000000,0.000000,1.000000}%
\pgfsetstrokecolor{currentstroke}%
\pgfsetdash{{11.100000pt}{4.800000pt}}{0.000000pt}%
\pgfpathmoveto{\pgfqpoint{2.612960in}{8.072852in}}%
\pgfpathlineto{\pgfqpoint{2.734066in}{6.982852in}}%
\pgfpathlineto{\pgfqpoint{2.855172in}{4.362158in}}%
\pgfpathlineto{\pgfqpoint{2.976278in}{4.155843in}}%
\pgfpathlineto{\pgfqpoint{3.097384in}{4.064122in}}%
\pgfpathlineto{\pgfqpoint{3.218490in}{3.511507in}}%
\pgfpathlineto{\pgfqpoint{3.339596in}{3.351655in}}%
\pgfpathlineto{\pgfqpoint{3.460702in}{3.340666in}}%
\pgfpathlineto{\pgfqpoint{3.581808in}{3.323468in}}%
\pgfpathlineto{\pgfqpoint{3.702914in}{3.308755in}}%
\pgfpathlineto{\pgfqpoint{3.824020in}{3.289404in}}%
\pgfpathlineto{\pgfqpoint{3.945126in}{3.281454in}}%
\pgfpathlineto{\pgfqpoint{4.066232in}{3.265508in}}%
\pgfpathlineto{\pgfqpoint{4.187338in}{3.262320in}}%
\pgfpathlineto{\pgfqpoint{4.308444in}{3.266267in}}%
\pgfpathlineto{\pgfqpoint{4.429550in}{3.268995in}}%
\pgfpathlineto{\pgfqpoint{4.550656in}{3.285808in}}%
\pgfpathlineto{\pgfqpoint{4.671762in}{3.286900in}}%
\pgfpathlineto{\pgfqpoint{4.792868in}{3.304449in}}%
\pgfpathlineto{\pgfqpoint{4.913974in}{3.315484in}}%
\pgfpathlineto{\pgfqpoint{5.035080in}{3.324204in}}%
\pgfpathlineto{\pgfqpoint{5.156186in}{3.331304in}}%
\pgfpathlineto{\pgfqpoint{5.277292in}{3.332051in}}%
\pgfpathlineto{\pgfqpoint{5.398398in}{3.331144in}}%
\pgfpathlineto{\pgfqpoint{5.519504in}{3.336516in}}%
\pgfpathlineto{\pgfqpoint{5.640610in}{3.349377in}}%
\pgfpathlineto{\pgfqpoint{5.761716in}{3.361686in}}%
\pgfpathlineto{\pgfqpoint{5.882822in}{3.376019in}}%
\pgfpathlineto{\pgfqpoint{6.003928in}{3.379943in}}%
\pgfpathlineto{\pgfqpoint{6.125034in}{3.381310in}}%
\pgfpathlineto{\pgfqpoint{6.246140in}{3.381305in}}%
\pgfpathlineto{\pgfqpoint{6.367246in}{3.387320in}}%
\pgfpathlineto{\pgfqpoint{6.488352in}{3.406097in}}%
\pgfpathlineto{\pgfqpoint{6.609458in}{3.433942in}}%
\pgfpathlineto{\pgfqpoint{6.730564in}{3.454751in}}%
\pgfpathlineto{\pgfqpoint{6.851670in}{3.466909in}}%
\pgfpathlineto{\pgfqpoint{6.972776in}{3.479619in}}%
\pgfpathlineto{\pgfqpoint{7.093882in}{3.487280in}}%
\pgfpathlineto{\pgfqpoint{7.214988in}{3.491820in}}%
\pgfpathlineto{\pgfqpoint{7.336094in}{3.495227in}}%
\pgfpathlineto{\pgfqpoint{7.457200in}{3.497156in}}%
\pgfpathlineto{\pgfqpoint{7.578306in}{3.496480in}}%
\pgfpathlineto{\pgfqpoint{7.699412in}{3.493110in}}%
\pgfpathlineto{\pgfqpoint{7.820518in}{3.487822in}}%
\pgfpathlineto{\pgfqpoint{7.941624in}{3.482690in}}%
\pgfpathlineto{\pgfqpoint{8.062730in}{3.480727in}}%
\pgfpathlineto{\pgfqpoint{8.183836in}{3.482124in}}%
\pgfpathlineto{\pgfqpoint{8.304942in}{3.482994in}}%
\pgfpathlineto{\pgfqpoint{8.426048in}{3.483268in}}%
\pgfpathlineto{\pgfqpoint{8.547154in}{3.480445in}}%
\pgfusepath{stroke}%
\end{pgfscope}%
\begin{pgfscope}%
\pgfsetrectcap%
\pgfsetmiterjoin%
\pgfsetlinewidth{0.803000pt}%
\definecolor{currentstroke}{rgb}{0.000000,0.000000,0.000000}%
\pgfsetstrokecolor{currentstroke}%
\pgfsetdash{}{0pt}%
\pgfpathmoveto{\pgfqpoint{2.316250in}{1.090100in}}%
\pgfpathlineto{\pgfqpoint{2.316250in}{8.720800in}}%
\pgfusepath{stroke}%
\end{pgfscope}%
\begin{pgfscope}%
\pgfsetrectcap%
\pgfsetmiterjoin%
\pgfsetlinewidth{0.803000pt}%
\definecolor{currentstroke}{rgb}{0.000000,0.000000,0.000000}%
\pgfsetstrokecolor{currentstroke}%
\pgfsetdash{}{0pt}%
\pgfpathmoveto{\pgfqpoint{8.843864in}{1.090100in}}%
\pgfpathlineto{\pgfqpoint{8.843864in}{8.720800in}}%
\pgfusepath{stroke}%
\end{pgfscope}%
\begin{pgfscope}%
\pgfsetrectcap%
\pgfsetmiterjoin%
\pgfsetlinewidth{0.803000pt}%
\definecolor{currentstroke}{rgb}{0.000000,0.000000,0.000000}%
\pgfsetstrokecolor{currentstroke}%
\pgfsetdash{}{0pt}%
\pgfpathmoveto{\pgfqpoint{2.316250in}{1.090100in}}%
\pgfpathlineto{\pgfqpoint{8.843864in}{1.090100in}}%
\pgfusepath{stroke}%
\end{pgfscope}%
\begin{pgfscope}%
\pgfsetrectcap%
\pgfsetmiterjoin%
\pgfsetlinewidth{0.803000pt}%
\definecolor{currentstroke}{rgb}{0.000000,0.000000,0.000000}%
\pgfsetstrokecolor{currentstroke}%
\pgfsetdash{}{0pt}%
\pgfpathmoveto{\pgfqpoint{2.316250in}{8.720800in}}%
\pgfpathlineto{\pgfqpoint{8.843864in}{8.720800in}}%
\pgfusepath{stroke}%
\end{pgfscope}%
\begin{pgfscope}%
\pgfsetbuttcap%
\pgfsetmiterjoin%
\definecolor{currentfill}{rgb}{1.000000,1.000000,1.000000}%
\pgfsetfillcolor{currentfill}%
\pgfsetfillopacity{0.800000}%
\pgfsetlinewidth{1.003750pt}%
\definecolor{currentstroke}{rgb}{0.800000,0.800000,0.800000}%
\pgfsetstrokecolor{currentstroke}%
\pgfsetstrokeopacity{0.800000}%
\pgfsetdash{}{0pt}%
\pgfpathmoveto{\pgfqpoint{4.130341in}{7.901882in}}%
\pgfpathlineto{\pgfqpoint{8.591086in}{7.901882in}}%
\pgfpathquadraticcurveto{\pgfqpoint{8.663308in}{7.901882in}}{\pgfqpoint{8.663308in}{7.974104in}}%
\pgfpathlineto{\pgfqpoint{8.663308in}{8.468022in}}%
\pgfpathquadraticcurveto{\pgfqpoint{8.663308in}{8.540244in}}{\pgfqpoint{8.591086in}{8.540244in}}%
\pgfpathlineto{\pgfqpoint{4.130341in}{8.540244in}}%
\pgfpathquadraticcurveto{\pgfqpoint{4.058119in}{8.540244in}}{\pgfqpoint{4.058119in}{8.468022in}}%
\pgfpathlineto{\pgfqpoint{4.058119in}{7.974104in}}%
\pgfpathquadraticcurveto{\pgfqpoint{4.058119in}{7.901882in}}{\pgfqpoint{4.130341in}{7.901882in}}%
\pgfpathclose%
\pgfusepath{stroke,fill}%
\end{pgfscope}%
\begin{pgfscope}%
\pgfsetbuttcap%
\pgfsetroundjoin%
\pgfsetlinewidth{3.011250pt}%
\definecolor{currentstroke}{rgb}{0.000000,0.000000,1.000000}%
\pgfsetstrokecolor{currentstroke}%
\pgfsetdash{{11.100000pt}{4.800000pt}}{0.000000pt}%
\pgfpathmoveto{\pgfqpoint{4.202563in}{8.247829in}}%
\pgfpathlineto{\pgfqpoint{4.924785in}{8.247829in}}%
\pgfusepath{stroke}%
\end{pgfscope}%
\begin{pgfscope}%
\definecolor{textcolor}{rgb}{0.000000,0.000000,0.000000}%
\pgfsetstrokecolor{textcolor}%
\pgfsetfillcolor{textcolor}%
\pgftext[x=5.213674in,y=8.121440in,left,base]{\color{textcolor}\sffamily\fontsize{26.000000}{31.200000}\selectfont Seqtoseq Network}%
\end{pgfscope}%
\begin{pgfscope}%
\pgfsetbuttcap%
\pgfsetmiterjoin%
\definecolor{currentfill}{rgb}{1.000000,1.000000,1.000000}%
\pgfsetfillcolor{currentfill}%
\pgfsetlinewidth{0.000000pt}%
\definecolor{currentstroke}{rgb}{0.000000,0.000000,0.000000}%
\pgfsetstrokecolor{currentstroke}%
\pgfsetstrokeopacity{0.000000}%
\pgfsetdash{}{0pt}%
\pgfpathmoveto{\pgfqpoint{10.149386in}{1.090100in}}%
\pgfpathlineto{\pgfqpoint{16.677000in}{1.090100in}}%
\pgfpathlineto{\pgfqpoint{16.677000in}{8.720800in}}%
\pgfpathlineto{\pgfqpoint{10.149386in}{8.720800in}}%
\pgfpathclose%
\pgfusepath{fill}%
\end{pgfscope}%
\begin{pgfscope}%
\pgfsetbuttcap%
\pgfsetroundjoin%
\definecolor{currentfill}{rgb}{0.000000,0.000000,0.000000}%
\pgfsetfillcolor{currentfill}%
\pgfsetlinewidth{0.803000pt}%
\definecolor{currentstroke}{rgb}{0.000000,0.000000,0.000000}%
\pgfsetstrokecolor{currentstroke}%
\pgfsetdash{}{0pt}%
\pgfsys@defobject{currentmarker}{\pgfqpoint{0.000000in}{-0.048611in}}{\pgfqpoint{0.000000in}{0.000000in}}{%
\pgfpathmoveto{\pgfqpoint{0.000000in}{0.000000in}}%
\pgfpathlineto{\pgfqpoint{0.000000in}{-0.048611in}}%
\pgfusepath{stroke,fill}%
}%
\begin{pgfscope}%
\pgfsys@transformshift{10.446096in}{1.090100in}%
\pgfsys@useobject{currentmarker}{}%
\end{pgfscope}%
\end{pgfscope}%
\begin{pgfscope}%
\definecolor{textcolor}{rgb}{0.000000,0.000000,0.000000}%
\pgfsetstrokecolor{textcolor}%
\pgfsetfillcolor{textcolor}%
\pgftext[x=10.446096in,y=0.992878in,,top]{\color{textcolor}\sffamily\fontsize{20.000000}{24.000000}\selectfont 0}%
\end{pgfscope}%
\begin{pgfscope}%
\pgfsetbuttcap%
\pgfsetroundjoin%
\definecolor{currentfill}{rgb}{0.000000,0.000000,0.000000}%
\pgfsetfillcolor{currentfill}%
\pgfsetlinewidth{0.803000pt}%
\definecolor{currentstroke}{rgb}{0.000000,0.000000,0.000000}%
\pgfsetstrokecolor{currentstroke}%
\pgfsetdash{}{0pt}%
\pgfsys@defobject{currentmarker}{\pgfqpoint{0.000000in}{-0.048611in}}{\pgfqpoint{0.000000in}{0.000000in}}{%
\pgfpathmoveto{\pgfqpoint{0.000000in}{0.000000in}}%
\pgfpathlineto{\pgfqpoint{0.000000in}{-0.048611in}}%
\pgfusepath{stroke,fill}%
}%
\begin{pgfscope}%
\pgfsys@transformshift{11.657156in}{1.090100in}%
\pgfsys@useobject{currentmarker}{}%
\end{pgfscope}%
\end{pgfscope}%
\begin{pgfscope}%
\definecolor{textcolor}{rgb}{0.000000,0.000000,0.000000}%
\pgfsetstrokecolor{textcolor}%
\pgfsetfillcolor{textcolor}%
\pgftext[x=11.657156in,y=0.992878in,,top]{\color{textcolor}\sffamily\fontsize{20.000000}{24.000000}\selectfont 10}%
\end{pgfscope}%
\begin{pgfscope}%
\pgfsetbuttcap%
\pgfsetroundjoin%
\definecolor{currentfill}{rgb}{0.000000,0.000000,0.000000}%
\pgfsetfillcolor{currentfill}%
\pgfsetlinewidth{0.803000pt}%
\definecolor{currentstroke}{rgb}{0.000000,0.000000,0.000000}%
\pgfsetstrokecolor{currentstroke}%
\pgfsetdash{}{0pt}%
\pgfsys@defobject{currentmarker}{\pgfqpoint{0.000000in}{-0.048611in}}{\pgfqpoint{0.000000in}{0.000000in}}{%
\pgfpathmoveto{\pgfqpoint{0.000000in}{0.000000in}}%
\pgfpathlineto{\pgfqpoint{0.000000in}{-0.048611in}}%
\pgfusepath{stroke,fill}%
}%
\begin{pgfscope}%
\pgfsys@transformshift{12.868216in}{1.090100in}%
\pgfsys@useobject{currentmarker}{}%
\end{pgfscope}%
\end{pgfscope}%
\begin{pgfscope}%
\definecolor{textcolor}{rgb}{0.000000,0.000000,0.000000}%
\pgfsetstrokecolor{textcolor}%
\pgfsetfillcolor{textcolor}%
\pgftext[x=12.868216in,y=0.992878in,,top]{\color{textcolor}\sffamily\fontsize{20.000000}{24.000000}\selectfont 20}%
\end{pgfscope}%
\begin{pgfscope}%
\pgfsetbuttcap%
\pgfsetroundjoin%
\definecolor{currentfill}{rgb}{0.000000,0.000000,0.000000}%
\pgfsetfillcolor{currentfill}%
\pgfsetlinewidth{0.803000pt}%
\definecolor{currentstroke}{rgb}{0.000000,0.000000,0.000000}%
\pgfsetstrokecolor{currentstroke}%
\pgfsetdash{}{0pt}%
\pgfsys@defobject{currentmarker}{\pgfqpoint{0.000000in}{-0.048611in}}{\pgfqpoint{0.000000in}{0.000000in}}{%
\pgfpathmoveto{\pgfqpoint{0.000000in}{0.000000in}}%
\pgfpathlineto{\pgfqpoint{0.000000in}{-0.048611in}}%
\pgfusepath{stroke,fill}%
}%
\begin{pgfscope}%
\pgfsys@transformshift{14.079276in}{1.090100in}%
\pgfsys@useobject{currentmarker}{}%
\end{pgfscope}%
\end{pgfscope}%
\begin{pgfscope}%
\definecolor{textcolor}{rgb}{0.000000,0.000000,0.000000}%
\pgfsetstrokecolor{textcolor}%
\pgfsetfillcolor{textcolor}%
\pgftext[x=14.079276in,y=0.992878in,,top]{\color{textcolor}\sffamily\fontsize{20.000000}{24.000000}\selectfont 30}%
\end{pgfscope}%
\begin{pgfscope}%
\pgfsetbuttcap%
\pgfsetroundjoin%
\definecolor{currentfill}{rgb}{0.000000,0.000000,0.000000}%
\pgfsetfillcolor{currentfill}%
\pgfsetlinewidth{0.803000pt}%
\definecolor{currentstroke}{rgb}{0.000000,0.000000,0.000000}%
\pgfsetstrokecolor{currentstroke}%
\pgfsetdash{}{0pt}%
\pgfsys@defobject{currentmarker}{\pgfqpoint{0.000000in}{-0.048611in}}{\pgfqpoint{0.000000in}{0.000000in}}{%
\pgfpathmoveto{\pgfqpoint{0.000000in}{0.000000in}}%
\pgfpathlineto{\pgfqpoint{0.000000in}{-0.048611in}}%
\pgfusepath{stroke,fill}%
}%
\begin{pgfscope}%
\pgfsys@transformshift{15.290336in}{1.090100in}%
\pgfsys@useobject{currentmarker}{}%
\end{pgfscope}%
\end{pgfscope}%
\begin{pgfscope}%
\definecolor{textcolor}{rgb}{0.000000,0.000000,0.000000}%
\pgfsetstrokecolor{textcolor}%
\pgfsetfillcolor{textcolor}%
\pgftext[x=15.290336in,y=0.992878in,,top]{\color{textcolor}\sffamily\fontsize{20.000000}{24.000000}\selectfont 40}%
\end{pgfscope}%
\begin{pgfscope}%
\pgfsetbuttcap%
\pgfsetroundjoin%
\definecolor{currentfill}{rgb}{0.000000,0.000000,0.000000}%
\pgfsetfillcolor{currentfill}%
\pgfsetlinewidth{0.803000pt}%
\definecolor{currentstroke}{rgb}{0.000000,0.000000,0.000000}%
\pgfsetstrokecolor{currentstroke}%
\pgfsetdash{}{0pt}%
\pgfsys@defobject{currentmarker}{\pgfqpoint{0.000000in}{-0.048611in}}{\pgfqpoint{0.000000in}{0.000000in}}{%
\pgfpathmoveto{\pgfqpoint{0.000000in}{0.000000in}}%
\pgfpathlineto{\pgfqpoint{0.000000in}{-0.048611in}}%
\pgfusepath{stroke,fill}%
}%
\begin{pgfscope}%
\pgfsys@transformshift{16.501396in}{1.090100in}%
\pgfsys@useobject{currentmarker}{}%
\end{pgfscope}%
\end{pgfscope}%
\begin{pgfscope}%
\definecolor{textcolor}{rgb}{0.000000,0.000000,0.000000}%
\pgfsetstrokecolor{textcolor}%
\pgfsetfillcolor{textcolor}%
\pgftext[x=16.501396in,y=0.992878in,,top]{\color{textcolor}\sffamily\fontsize{20.000000}{24.000000}\selectfont 50}%
\end{pgfscope}%
\begin{pgfscope}%
\definecolor{textcolor}{rgb}{0.000000,0.000000,0.000000}%
\pgfsetstrokecolor{textcolor}%
\pgfsetfillcolor{textcolor}%
\pgftext[x=13.413193in,y=0.668497in,,top]{\color{textcolor}\sffamily\fontsize{26.000000}{31.200000}\selectfont Training epochs}%
\end{pgfscope}%
\begin{pgfscope}%
\pgfsetbuttcap%
\pgfsetroundjoin%
\definecolor{currentfill}{rgb}{0.000000,0.000000,0.000000}%
\pgfsetfillcolor{currentfill}%
\pgfsetlinewidth{0.803000pt}%
\definecolor{currentstroke}{rgb}{0.000000,0.000000,0.000000}%
\pgfsetstrokecolor{currentstroke}%
\pgfsetdash{}{0pt}%
\pgfsys@defobject{currentmarker}{\pgfqpoint{-0.048611in}{0.000000in}}{\pgfqpoint{0.000000in}{0.000000in}}{%
\pgfpathmoveto{\pgfqpoint{0.000000in}{0.000000in}}%
\pgfpathlineto{\pgfqpoint{-0.048611in}{0.000000in}}%
\pgfusepath{stroke,fill}%
}%
\begin{pgfscope}%
\pgfsys@transformshift{10.149386in}{1.090100in}%
\pgfsys@useobject{currentmarker}{}%
\end{pgfscope}%
\end{pgfscope}%
\begin{pgfscope}%
\definecolor{textcolor}{rgb}{0.000000,0.000000,0.000000}%
\pgfsetstrokecolor{textcolor}%
\pgfsetfillcolor{textcolor}%
\pgftext[x=9.610405in,y=0.984577in,left,base]{\color{textcolor}\sffamily\fontsize{20.000000}{24.000000}\selectfont 0.2}%
\end{pgfscope}%
\begin{pgfscope}%
\pgfsetbuttcap%
\pgfsetroundjoin%
\definecolor{currentfill}{rgb}{0.000000,0.000000,0.000000}%
\pgfsetfillcolor{currentfill}%
\pgfsetlinewidth{0.803000pt}%
\definecolor{currentstroke}{rgb}{0.000000,0.000000,0.000000}%
\pgfsetstrokecolor{currentstroke}%
\pgfsetdash{}{0pt}%
\pgfsys@defobject{currentmarker}{\pgfqpoint{-0.048611in}{0.000000in}}{\pgfqpoint{0.000000in}{0.000000in}}{%
\pgfpathmoveto{\pgfqpoint{0.000000in}{0.000000in}}%
\pgfpathlineto{\pgfqpoint{-0.048611in}{0.000000in}}%
\pgfusepath{stroke,fill}%
}%
\begin{pgfscope}%
\pgfsys@transformshift{10.149386in}{2.616240in}%
\pgfsys@useobject{currentmarker}{}%
\end{pgfscope}%
\end{pgfscope}%
\begin{pgfscope}%
\definecolor{textcolor}{rgb}{0.000000,0.000000,0.000000}%
\pgfsetstrokecolor{textcolor}%
\pgfsetfillcolor{textcolor}%
\pgftext[x=9.610405in,y=2.510717in,left,base]{\color{textcolor}\sffamily\fontsize{20.000000}{24.000000}\selectfont 0.3}%
\end{pgfscope}%
\begin{pgfscope}%
\pgfsetbuttcap%
\pgfsetroundjoin%
\definecolor{currentfill}{rgb}{0.000000,0.000000,0.000000}%
\pgfsetfillcolor{currentfill}%
\pgfsetlinewidth{0.803000pt}%
\definecolor{currentstroke}{rgb}{0.000000,0.000000,0.000000}%
\pgfsetstrokecolor{currentstroke}%
\pgfsetdash{}{0pt}%
\pgfsys@defobject{currentmarker}{\pgfqpoint{-0.048611in}{0.000000in}}{\pgfqpoint{0.000000in}{0.000000in}}{%
\pgfpathmoveto{\pgfqpoint{0.000000in}{0.000000in}}%
\pgfpathlineto{\pgfqpoint{-0.048611in}{0.000000in}}%
\pgfusepath{stroke,fill}%
}%
\begin{pgfscope}%
\pgfsys@transformshift{10.149386in}{4.142380in}%
\pgfsys@useobject{currentmarker}{}%
\end{pgfscope}%
\end{pgfscope}%
\begin{pgfscope}%
\definecolor{textcolor}{rgb}{0.000000,0.000000,0.000000}%
\pgfsetstrokecolor{textcolor}%
\pgfsetfillcolor{textcolor}%
\pgftext[x=9.610405in,y=4.036857in,left,base]{\color{textcolor}\sffamily\fontsize{20.000000}{24.000000}\selectfont 0.4}%
\end{pgfscope}%
\begin{pgfscope}%
\pgfsetbuttcap%
\pgfsetroundjoin%
\definecolor{currentfill}{rgb}{0.000000,0.000000,0.000000}%
\pgfsetfillcolor{currentfill}%
\pgfsetlinewidth{0.803000pt}%
\definecolor{currentstroke}{rgb}{0.000000,0.000000,0.000000}%
\pgfsetstrokecolor{currentstroke}%
\pgfsetdash{}{0pt}%
\pgfsys@defobject{currentmarker}{\pgfqpoint{-0.048611in}{0.000000in}}{\pgfqpoint{0.000000in}{0.000000in}}{%
\pgfpathmoveto{\pgfqpoint{0.000000in}{0.000000in}}%
\pgfpathlineto{\pgfqpoint{-0.048611in}{0.000000in}}%
\pgfusepath{stroke,fill}%
}%
\begin{pgfscope}%
\pgfsys@transformshift{10.149386in}{5.668520in}%
\pgfsys@useobject{currentmarker}{}%
\end{pgfscope}%
\end{pgfscope}%
\begin{pgfscope}%
\definecolor{textcolor}{rgb}{0.000000,0.000000,0.000000}%
\pgfsetstrokecolor{textcolor}%
\pgfsetfillcolor{textcolor}%
\pgftext[x=9.610405in,y=5.562997in,left,base]{\color{textcolor}\sffamily\fontsize{20.000000}{24.000000}\selectfont 0.5}%
\end{pgfscope}%
\begin{pgfscope}%
\pgfsetbuttcap%
\pgfsetroundjoin%
\definecolor{currentfill}{rgb}{0.000000,0.000000,0.000000}%
\pgfsetfillcolor{currentfill}%
\pgfsetlinewidth{0.803000pt}%
\definecolor{currentstroke}{rgb}{0.000000,0.000000,0.000000}%
\pgfsetstrokecolor{currentstroke}%
\pgfsetdash{}{0pt}%
\pgfsys@defobject{currentmarker}{\pgfqpoint{-0.048611in}{0.000000in}}{\pgfqpoint{0.000000in}{0.000000in}}{%
\pgfpathmoveto{\pgfqpoint{0.000000in}{0.000000in}}%
\pgfpathlineto{\pgfqpoint{-0.048611in}{0.000000in}}%
\pgfusepath{stroke,fill}%
}%
\begin{pgfscope}%
\pgfsys@transformshift{10.149386in}{7.194660in}%
\pgfsys@useobject{currentmarker}{}%
\end{pgfscope}%
\end{pgfscope}%
\begin{pgfscope}%
\definecolor{textcolor}{rgb}{0.000000,0.000000,0.000000}%
\pgfsetstrokecolor{textcolor}%
\pgfsetfillcolor{textcolor}%
\pgftext[x=9.610405in,y=7.089137in,left,base]{\color{textcolor}\sffamily\fontsize{20.000000}{24.000000}\selectfont 0.6}%
\end{pgfscope}%
\begin{pgfscope}%
\pgfsetbuttcap%
\pgfsetroundjoin%
\definecolor{currentfill}{rgb}{0.000000,0.000000,0.000000}%
\pgfsetfillcolor{currentfill}%
\pgfsetlinewidth{0.803000pt}%
\definecolor{currentstroke}{rgb}{0.000000,0.000000,0.000000}%
\pgfsetstrokecolor{currentstroke}%
\pgfsetdash{}{0pt}%
\pgfsys@defobject{currentmarker}{\pgfqpoint{-0.048611in}{0.000000in}}{\pgfqpoint{0.000000in}{0.000000in}}{%
\pgfpathmoveto{\pgfqpoint{0.000000in}{0.000000in}}%
\pgfpathlineto{\pgfqpoint{-0.048611in}{0.000000in}}%
\pgfusepath{stroke,fill}%
}%
\begin{pgfscope}%
\pgfsys@transformshift{10.149386in}{8.720800in}%
\pgfsys@useobject{currentmarker}{}%
\end{pgfscope}%
\end{pgfscope}%
\begin{pgfscope}%
\definecolor{textcolor}{rgb}{0.000000,0.000000,0.000000}%
\pgfsetstrokecolor{textcolor}%
\pgfsetfillcolor{textcolor}%
\pgftext[x=9.610405in,y=8.615277in,left,base]{\color{textcolor}\sffamily\fontsize{20.000000}{24.000000}\selectfont 0.7}%
\end{pgfscope}%
\begin{pgfscope}%
\pgfpathrectangle{\pgfqpoint{10.149386in}{1.090100in}}{\pgfqpoint{6.527614in}{7.630700in}}%
\pgfusepath{clip}%
\pgfsetbuttcap%
\pgfsetroundjoin%
\pgfsetlinewidth{3.011250pt}%
\definecolor{currentstroke}{rgb}{0.000000,0.500000,0.000000}%
\pgfsetstrokecolor{currentstroke}%
\pgfsetdash{{11.100000pt}{4.800000pt}}{0.000000pt}%
\pgfpathmoveto{\pgfqpoint{10.446096in}{7.686554in}}%
\pgfpathlineto{\pgfqpoint{10.567202in}{7.006248in}}%
\pgfpathlineto{\pgfqpoint{10.688308in}{3.277454in}}%
\pgfpathlineto{\pgfqpoint{10.809414in}{2.618389in}}%
\pgfpathlineto{\pgfqpoint{10.930520in}{2.611620in}}%
\pgfpathlineto{\pgfqpoint{11.051626in}{2.452816in}}%
\pgfpathlineto{\pgfqpoint{11.172732in}{2.292390in}}%
\pgfpathlineto{\pgfqpoint{11.293838in}{2.278712in}}%
\pgfpathlineto{\pgfqpoint{11.414944in}{2.274838in}}%
\pgfpathlineto{\pgfqpoint{11.536050in}{2.267503in}}%
\pgfpathlineto{\pgfqpoint{11.657156in}{2.259714in}}%
\pgfpathlineto{\pgfqpoint{11.778262in}{2.258001in}}%
\pgfpathlineto{\pgfqpoint{11.899368in}{2.253457in}}%
\pgfpathlineto{\pgfqpoint{12.020474in}{2.245817in}}%
\pgfpathlineto{\pgfqpoint{12.141580in}{2.241856in}}%
\pgfpathlineto{\pgfqpoint{12.262686in}{2.235588in}}%
\pgfpathlineto{\pgfqpoint{12.383792in}{2.231325in}}%
\pgfpathlineto{\pgfqpoint{12.504898in}{2.224969in}}%
\pgfpathlineto{\pgfqpoint{12.626004in}{2.220294in}}%
\pgfpathlineto{\pgfqpoint{12.747110in}{2.224842in}}%
\pgfpathlineto{\pgfqpoint{12.868216in}{2.222913in}}%
\pgfpathlineto{\pgfqpoint{12.989322in}{2.221487in}}%
\pgfpathlineto{\pgfqpoint{13.110428in}{2.225776in}}%
\pgfpathlineto{\pgfqpoint{13.231534in}{2.215116in}}%
\pgfpathlineto{\pgfqpoint{13.352640in}{2.209811in}}%
\pgfpathlineto{\pgfqpoint{13.473746in}{2.209769in}}%
\pgfpathlineto{\pgfqpoint{13.594852in}{2.207845in}}%
\pgfpathlineto{\pgfqpoint{13.715958in}{2.204234in}}%
\pgfpathlineto{\pgfqpoint{13.837064in}{2.196342in}}%
\pgfpathlineto{\pgfqpoint{13.958170in}{2.195762in}}%
\pgfpathlineto{\pgfqpoint{14.079276in}{2.195980in}}%
\pgfpathlineto{\pgfqpoint{14.200382in}{2.200619in}}%
\pgfpathlineto{\pgfqpoint{14.321488in}{2.197533in}}%
\pgfpathlineto{\pgfqpoint{14.442594in}{2.198128in}}%
\pgfpathlineto{\pgfqpoint{14.563700in}{2.203622in}}%
\pgfpathlineto{\pgfqpoint{14.684806in}{2.208453in}}%
\pgfpathlineto{\pgfqpoint{14.805912in}{2.214028in}}%
\pgfpathlineto{\pgfqpoint{14.927018in}{2.221425in}}%
\pgfpathlineto{\pgfqpoint{15.048124in}{2.228624in}}%
\pgfpathlineto{\pgfqpoint{15.169230in}{2.226751in}}%
\pgfpathlineto{\pgfqpoint{15.290336in}{2.237818in}}%
\pgfpathlineto{\pgfqpoint{15.411442in}{2.235482in}}%
\pgfpathlineto{\pgfqpoint{15.532548in}{2.248417in}}%
\pgfpathlineto{\pgfqpoint{15.653654in}{2.242851in}}%
\pgfpathlineto{\pgfqpoint{15.774760in}{2.246614in}}%
\pgfpathlineto{\pgfqpoint{15.895866in}{2.249952in}}%
\pgfpathlineto{\pgfqpoint{16.016972in}{2.253660in}}%
\pgfpathlineto{\pgfqpoint{16.138078in}{2.259135in}}%
\pgfpathlineto{\pgfqpoint{16.259184in}{2.259684in}}%
\pgfpathlineto{\pgfqpoint{16.380290in}{2.262490in}}%
\pgfusepath{stroke}%
\end{pgfscope}%
\begin{pgfscope}%
\pgfsetrectcap%
\pgfsetmiterjoin%
\pgfsetlinewidth{0.803000pt}%
\definecolor{currentstroke}{rgb}{0.000000,0.000000,0.000000}%
\pgfsetstrokecolor{currentstroke}%
\pgfsetdash{}{0pt}%
\pgfpathmoveto{\pgfqpoint{10.149386in}{1.090100in}}%
\pgfpathlineto{\pgfqpoint{10.149386in}{8.720800in}}%
\pgfusepath{stroke}%
\end{pgfscope}%
\begin{pgfscope}%
\pgfsetrectcap%
\pgfsetmiterjoin%
\pgfsetlinewidth{0.803000pt}%
\definecolor{currentstroke}{rgb}{0.000000,0.000000,0.000000}%
\pgfsetstrokecolor{currentstroke}%
\pgfsetdash{}{0pt}%
\pgfpathmoveto{\pgfqpoint{16.677000in}{1.090100in}}%
\pgfpathlineto{\pgfqpoint{16.677000in}{8.720800in}}%
\pgfusepath{stroke}%
\end{pgfscope}%
\begin{pgfscope}%
\pgfsetrectcap%
\pgfsetmiterjoin%
\pgfsetlinewidth{0.803000pt}%
\definecolor{currentstroke}{rgb}{0.000000,0.000000,0.000000}%
\pgfsetstrokecolor{currentstroke}%
\pgfsetdash{}{0pt}%
\pgfpathmoveto{\pgfqpoint{10.149386in}{1.090100in}}%
\pgfpathlineto{\pgfqpoint{16.677000in}{1.090100in}}%
\pgfusepath{stroke}%
\end{pgfscope}%
\begin{pgfscope}%
\pgfsetrectcap%
\pgfsetmiterjoin%
\pgfsetlinewidth{0.803000pt}%
\definecolor{currentstroke}{rgb}{0.000000,0.000000,0.000000}%
\pgfsetstrokecolor{currentstroke}%
\pgfsetdash{}{0pt}%
\pgfpathmoveto{\pgfqpoint{10.149386in}{8.720800in}}%
\pgfpathlineto{\pgfqpoint{16.677000in}{8.720800in}}%
\pgfusepath{stroke}%
\end{pgfscope}%
\begin{pgfscope}%
\pgfsetbuttcap%
\pgfsetmiterjoin%
\definecolor{currentfill}{rgb}{1.000000,1.000000,1.000000}%
\pgfsetfillcolor{currentfill}%
\pgfsetfillopacity{0.800000}%
\pgfsetlinewidth{1.003750pt}%
\definecolor{currentstroke}{rgb}{0.800000,0.800000,0.800000}%
\pgfsetstrokecolor{currentstroke}%
\pgfsetstrokeopacity{0.800000}%
\pgfsetdash{}{0pt}%
\pgfpathmoveto{\pgfqpoint{11.719974in}{7.901882in}}%
\pgfpathlineto{\pgfqpoint{16.424222in}{7.901882in}}%
\pgfpathquadraticcurveto{\pgfqpoint{16.496444in}{7.901882in}}{\pgfqpoint{16.496444in}{7.974104in}}%
\pgfpathlineto{\pgfqpoint{16.496444in}{8.468022in}}%
\pgfpathquadraticcurveto{\pgfqpoint{16.496444in}{8.540244in}}{\pgfqpoint{16.424222in}{8.540244in}}%
\pgfpathlineto{\pgfqpoint{11.719974in}{8.540244in}}%
\pgfpathquadraticcurveto{\pgfqpoint{11.647752in}{8.540244in}}{\pgfqpoint{11.647752in}{8.468022in}}%
\pgfpathlineto{\pgfqpoint{11.647752in}{7.974104in}}%
\pgfpathquadraticcurveto{\pgfqpoint{11.647752in}{7.901882in}}{\pgfqpoint{11.719974in}{7.901882in}}%
\pgfpathclose%
\pgfusepath{stroke,fill}%
\end{pgfscope}%
\begin{pgfscope}%
\pgfsetbuttcap%
\pgfsetroundjoin%
\pgfsetlinewidth{3.011250pt}%
\definecolor{currentstroke}{rgb}{0.000000,0.500000,0.000000}%
\pgfsetstrokecolor{currentstroke}%
\pgfsetdash{{11.100000pt}{4.800000pt}}{0.000000pt}%
\pgfpathmoveto{\pgfqpoint{11.792196in}{8.247829in}}%
\pgfpathlineto{\pgfqpoint{12.514419in}{8.247829in}}%
\pgfusepath{stroke}%
\end{pgfscope}%
\begin{pgfscope}%
\definecolor{textcolor}{rgb}{0.000000,0.000000,0.000000}%
\pgfsetstrokecolor{textcolor}%
\pgfsetfillcolor{textcolor}%
\pgftext[x=12.803307in,y=8.121440in,left,base]{\color{textcolor}\sffamily\fontsize{26.000000}{31.200000}\selectfont LSTM+FCL Network}%
\end{pgfscope}%
\end{pgfpicture}%
\makeatother%
\endgroup%